\documentclass[runningheads]{llncs}
\usepackage[T1]{fontenc}
\usepackage{graphicx}
\usepackage[utf8]{inputenc}
\usepackage{subcaption}
\usepackage{amsmath}
\usepackage{amssymb}
\usepackage{booktabs}
\usepackage{bbm}
\usepackage{algorithm,algpseudocode}
\usepackage[export]{adjustbox}
\algnewcommand{\Inputs}[1]{%
  \State \textbf{Inputs:}
  \Statex \hspace*{\algorithmicindent}\parbox[t]{.8\linewidth}{\raggedright #1}
}
\algnewcommand{\Outputs}[1]{%
  \State \textbf{Outputs:}
  \Statex \hspace*{\algorithmicindent}\parbox[t]{.8\linewidth}{\raggedright #1}
}
\algnewcommand{\Initialize}[1]{%
  \State \textbf{Initialize:}
  \Statex \hspace*{\algorithmicindent}\parbox[t]{.8\linewidth}{\raggedright #1}
}
\usepackage{algpseudocode}
\usepackage{amsmath}
\usepackage[pagebackref,breaklinks,colorlinks]{hyperref}
\usepackage{multirow}
\usepackage{array}
\newcolumntype{P}[1]{>{\centering\arraybackslash}p{#1}}

\usepackage[capitalize]{cleveref}
\crefname{section}{Sec.}{Secs.}
\Crefname{section}{Section}{Sections}
\Crefname{table}{Table}{Tables}
\crefname{table}{Tab.}{Tabs.}

\newcommand{\columnname}[1]
{\makebox[0.15\textwidth][c]{\textbf{#1}}}

\usepackage{color}

\begin{document}
\title{Graph Laplacian for Semi-Supervised Learning}
%
%

\author{Or Streicher \and Guy Gilboa}

\authorrunning{Or Streicher \and Guy Gilboa}


\institute{Technion - Israel Institute of Technology, Haifa 32000, Israel \\
\email{orr.shtr@gmail.com, guy.gilboa@ee.technion.ac.il}}

\maketitle              
\begin{abstract}
Semi-supervised learning is highly useful in common scenarios where labeled data is scarce but unlabeled data is abundant. 
The graph (or nonlocal) Laplacian is a fundamental smoothing operator for solving various learning tasks. For unsupervised clustering, a spectral embedding is often used, based on graph-Laplacian eigenvectors. For semi-supervised problems, the common approach is to solve a constrained optimization problem, regularized by a Dirichlet energy, based on the graph-Laplacian. 
However, as supervision decreases,  Dirichlet optimization becomes suboptimal. We therefore would like to obtain a smooth transition between unsupervised clustering and low-supervised graph-based classification. 

In this paper, we propose a new type of graph-Laplacian which is adapted for \emph{Semi-Supervised Learning} (SSL) problems.
It is based on both density and contrastive measures and allows the encoding of the labeled data directly in the operator.
Thus, we can perform successfully semi-supervised learning using spectral clustering. 
The benefits of our approach are illustrated for several SSL problems. 
\keywords{Graph Representation \and Semi-Supervise Learning  \and Nonlocal Laplacian \and Spectral Clustering}
\end{abstract}
\section{Introduction}
Labeling information is a major challenge in modern learning techniques, which in many cases can be a long and expensive process. A possible solution to this problem is Semi-Supervised Learning (SSL). SSL methods can be thought of as the halfway between supervised and unsupervised learning. It uses large amounts of unlabeled data and a limited amount of labeled data, to improve the learning model. SSL techniques are usually used when one cannot employ supervised learning algorithms. The use of supervised learning when limited labels are available may result in a lack of generalization and overfitting. Those limited known labels, however, can significantly improve performance compared to unsupervised learning algorithms \cite{hearty2016advanced}. Intuitively, the purpose of SSL is to \emph{generalize} the known labels to the unlabeled samples by an appropriate smoothing operator. The graph-Laplacian has shown to be highly effective for this purpose.

In this paper, we focus on
\emph{Graph-based methods} which are well-studied classical techniques. We note that the insights presented here can be further used by deep learning methods with spectral-graph modules, as in \cite{shaham2018spectralnet}, \cite{chen2022specnet2}, \cite{streicher2022basis}, \cite{aviles2022graphxcovid}. 
In classical graph methods, a weighted graph is constructed based on the affinities between data instances. These affinities are usually computed using a metric of the features representing each instance.
The vast majority of graph-based learning methods use the graph-Laplacian as the smoothing operator for generalization.
More advanced nonlinear methods use $p$-Laplacian operators \cite{elmoataz2017game}, \cite{calder2018game}.
In this study we limit the scope to the linear case (or quadratic energy), noting that our proposed operators can be further generalized.
Data processing based on the Laplacian has shown to be effective for a wide range of problems including clustering \cite{bresson2013multiclass} \cite{ng2001spectral}, \cite{zelnik2004self}, classification \cite{garcia2014multiclass},  segmentation \cite{shi2000normalized}, dimensionality reduction \cite{belkin2003laplacian}, \cite{coifman2006diffusion}, \cite{roweis2000nonlinear} and more.
For the SSL setting, most graph-based learning techniques use the properties of the graph-Laplacian operator to define an optimization problem. The labeled information can be inserted as problem constraints, see e.g. \cite{joachims2003transductive}, \cite{belkin2006manifold}, \cite{garcia2014multiclass}, \cite{mao2012parameter}, \cite{liu2010large}, \cite{shi2017weighted}. 

In this paper, we propose a different approach to consider the labeled information, by inserting it into the affinity measure that defines the connectivity between the nodes of the graph.
We first examine the work of \cite{shi2017weighted} on the weighted nonlocal Laplacian. In this work, the density of the labeled data is essentially increased. This improves the solution of the constrained optimization problem. We found that it has a marginal effect on the spectral embedding. Based on contrastive arguments, we propose to increase connections between labeled and unlabeled data and to increase (remove) connections between labeled data of the same (different) clusters. This yields a considerably improved spectral embedding. 

Our proposed method retains the following main qualities: 1) {\bf Interpolation between unsupervised and semi-supervised learning.} The suggested approach enables learning for a changing range of labeled information. 
2) {\bf Low-label regime performance.} The proposed method was found most advantageous in the low-label regime, compared to competitive techniques. 
3) {\bf Different analysis tools.} Wide variety of analysis tools can be used for solving SSL problems, including spectral and functional analysis.
We illustrate the advantages of using this new definition on toy examples and on real data sets.

\section{Setting and Notation}
Let $X=\{x_i\}_{i=1}^n$ be a finite set of instances in $\mathbb{R}^d$.
These instances are represented as nodes on an undirected weighted graph $G = (V, E, W)$, where $V$ is the vertices set, $E$ is the edges set and $W$ is the adjacency matrix. The adjacency matrix is symmetric and is usually defined by a distance measure between the nodes. For example, a common choice is a Gaussian kernel with Euclidean distance,
\begin{equation}\label{eq:Gaussian kernel}
W_{ij}=\exp\left(-\frac{\|x_i-x_j\|_2^2}{2\sigma^2}\right),
\end{equation}
where $\sigma$ is a soft-threshold parameter.

The degree matrix $D$ is a diagonal matrix where $D_{ii}$ is the degree of the $i$-th vertex, i.e.,
\begin{equation}\label{eq:DegreeMat_def}
D_{ii}=\sum_{j}{W_{ij}}. 
\end{equation}
The graph-Laplacian operator is defined by,
\begin{equation}\label{Laplacian_def}
L := D-W .
\end{equation}
The graph-Laplacian is a symmetric, positive semi-definite matrix, i.e., $\forall f\in \mathbb{R}^n \text{ , } f^TLf \ge 0$.
For each vector $f\in \mathbb{R}^n$ it holds that 
\begin{equation}\label{L operator}
Lf \in \mathbb{R}^n \text{   ,   }
Lf(j)=\sum_{i=1}^n{W_{ij}(f_j-f_i)},
\end{equation}

\begin{equation}\label{L quadratic form}
f^TLf=\sum_{i=1}^n{\sum_{j=1}^n{W_{ij}(f_i-f_j)^2}}.
\end{equation}
The eigenvalues of $L$ are real and non-negative and sorted in ascending order $\lambda_1 \leq \lambda_2 \leq ... \leq \lambda_n$. 
The corresponding eigenvectors form an orthogonal basis,  denoted by $u_1,u_2...,u_n$. 
The sample $x_i$ can be represented in the spectral embedding space as the $i$th row of the matrix $U=\begin{bmatrix}
    u_1&\cdots&u_K
\end{bmatrix} \in \mathbb{R}^{n \times K}$, denoted as $\varphi_i$. More formally, the spectral embedding of an instance $x_i$  can be formulated as
\begin{equation}\label{embedding_def}
x_i \longmapsto \varphi_i = [u_1(i), u_2(i),..., u_K(i)]\in \mathbb{R}^K,
\end{equation}
where in most cases $K\ll d$.

For the SSL setting, let us define a discrete function $f \in \mathbb{R}^n$  over $X$ and $S \subset X$, such that $|S|= m,  m \leq n$,  be a subset of $X$ on which the values of $f$ are known, i.e., $f(x)=g(x), \forall x\in S $, for a given function $g$. 
The purpose of the SSL model is to find the values of $f$ of all data-points in $X$ constrained by the values of the set $S$.

The main SSL problem this work focuses on is clustering. To evaluate the clustering performance we examined two common measures. The first one is \emph{Normalized mutual information} (NMI) which is defined as, \begin{equation}\label{NMI_defenition}
NMI(c, \hat{c}) = \frac{I(c, \hat{c})}{\max\{H(c), H(\hat{c})\}},
\end{equation}
where $I(c, \hat{c})$ is the mutual information between the true labels $c$ and the clustering result $\hat{c}$ and $H(\cdot)$ denotes entropy. 
We remind the definitions of entropy for a random variable $U$ with distribution $p_U$, $H(U):=E[-\log p_U]$, where $E$ is the expected value. For two random variables $U,V$ with conditional probability $p_{V|U}$ the conditional entropy of $V$ given $U$ is
$H(V|U):=E[-\log p_{V|U}]$. Mutual information measures the dependence between two random variable and admits the following identities $I(U,V) = H(U)-H(U|V)= H(V)-H(V|U)$.
It is non-negative and equals zero when $U$ and $V$ are independent.

The second measure is \emph{Unsupervised Clustering Accuracy} (ACC) which is defined as,
\begin{equation}\label{ACC_defenition}
ACC(c, \hat{c}) = \frac{1}{n}\max_{\pi \in \Pi}{\sum_{i=1}^n{\mathbbm{1}\{c_i=\pi(\hat{c}_i)\}}},
\end{equation}
where $\Pi$ is the set of possible permutations of the clustering results. To choose the optimal permutation $\pi$ we used the Kuhn-Munkres algorithm \cite{munkres1957algorithms}.
Both indicators are in the range $[0,1]$, where high values indicate a better correspondence between the clustering result and the true labels.

\section{Graph-Laplacian for SSL}
\subsection{Motivation}

A common approach to solve SSL problems, based on the graph-Laplacian (GL), is to solve the Dirichlet problem
\begin{equation} \label{eq:gl_functional}
\begin{aligned}
    \min_{f
    } \frac{1}{2}\sum_{x_i,x_j \in X}W_{ij}(f(x_i)-f(x_j))^2, \\ 
\end{aligned}
\end{equation}
\[\text{s.t. } f(x)=g(x) , \,\,x\in S.\]
A solution obeys 
\begin{equation}\label{eq:gl_sol}
Lf(x_i)=\sum_{x_j\in X}W_{ij}(f(x_i)-f(x_j))=0 , x_i \in X\setminus S 
\end{equation}
\[f(x)=g(x) , x \in S.\]\\
Note that \cref{eq:gl_sol} can be represented in matrix form. First, we define a constraints vector $ b \in \mathbb{R}^n$ and a mask $M \in \mathbb{R}^{n \times n} $ such that

\begin{equation}
b_i=
  \begin{cases} 
     g(x_i) & \text{if } x_i \in S \\ 
     0  & \text{if } x_i \in X\setminus S 
  \end{cases}
\text{  ,  }
  M_{ij}=
  \begin{cases} 
    \frac{1}{L_{ii}} & \text{if } x_i \in S \text{ and } i=j\\
     0 & \text{if } x_i \in S \text{ and } i \neq j\\ 
     1  & \text{if } x_i \in X\setminus S, \forall j 
  \end{cases}
  ,
\end{equation}
where $L_{ii}$ is the $i$-th element of the diagonal of $L$. Now,  \cref{eq:gl_sol} can be introduced in matrix from by
\begin{equation}\label{Graph Laplacian matrix form}
(M\circ L)f=b,
\end{equation}
where $\circ$ denotes element-wise multiplication.

A main problem with GL, as shown in \cite{shi2017weighted}, is that for a low sample rate of the labeled set, $|S|/|X|$, the solution is not continuous at the sample points. Thus the GL solution does not interpolate well the constraint values. 
In \cite{shi2017weighted} the authors suggested solving the discontinuity problem by using Weighted Nonlocal Laplacian (WNLL), assigning a greater weight to the labeled set $S$ compared to the unlabeled set $ X \setminus S $. Formally, the optimization problem of WNLL is given by

\begin{equation}\label{eq:wnll_optimization}
\min_{f}\sum_{x_i\in X\setminus S}\sum_{x_j\in X}W_{ij}(f(x_i)-f(x_j))^2+\mu\sum_{x_i\in S}\sum_{x_j\in X}W_{ij}(f(x_i)-f(x_j))^2
\end{equation}
\[\text{s.t. } f(x)=g(x) , x\in S,\]
where $\mu$ is a regularization parameter. It was suggested to set 
\begin{equation}
    \label{eq:mu}
    \mu = |X|/|S|, 
\end{equation}
the inverse of the sample rate. This can be interpreted as increasing the density (or measure) of the labeled instances.
The solution of \cref{eq:wnll_optimization} is given by solving the following linear system,
\begin{equation}\label{eq:wnll_sol}
\sum_{x_j\in X}(W_{ij}+W_{ji})(f(x_i)-f(x_j))+(\mu-1)\sum_{x_j\in S}W_{ji}(f(x_i)-f(x_j))=0 , x_i \in X\setminus S
\end{equation}
\[f(x)=g(x)  , x \in S.\]
Similarly to \cref{Graph Laplacian matrix form}, one can define \cref{eq:wnll_sol} in matrix form. Let us introduce the linear system as follows,
\begin{equation}\label{eq:wnll_sol_w_labeled}
\sum_{x_j\in X}(W_{ij}+W_{ji})(f(x_i)-f(x_j))+(\mu-1)\sum_{x_j\in X}W^{labeled}_{ij}(f(x_i)-f(x_j))=0 , x_i \in X\setminus S
\end{equation}
\[f(x)=g(x)  , x \in S,\]
such that, 
\begin{equation} \label{W_labeled_wnll}
  W^{labeled}_{ij}=
  \begin{cases} 
     W_{ij}  & x_i \in X \setminus S, x_j \in S \text{ or }  x_i \in  S, x_j \in X \setminus S \\
     0 & \textrm{otherwise}
  \end{cases}
\end{equation}
or equivalently,
\begin{equation}\label{eq:wnll_sol_sum_all}
\sum_{x_j\in X}\left(W_{ij}+W_{ji}+(\mu-1)W^{labeled}_{ij}\right)(f(x_i)-f(x_j))=0 , x_i \in X\setminus S
\end{equation}
\[f(x)=g(x)  , x \in S.\]
Now we can define,
\begin{equation}\label{eq:wnll_sol_sum_all_sol}
\sum_{x_j\in X}[W_{WNLL}]_{ji}(f(x_i)-f(x_j))=0 , x_i \in X\setminus S
\end{equation}
\[f(x)=g(x)  , x \in S,\]
where
\begin{equation}\label{W_ssl_for_wnll_definition}
W_{WNLL}= 2W+(\mu-1) W^{labeled}.
\end{equation}
Based on $W_{WNLL}$ one can define $L_{WNLL}$, following \cref{Laplacian_def}, 
such that  \cref{eq:wnll_sol} is equivalent to 
\begin{equation}\label{wnll_with_L_ssl }
(M \circ L_{WNLL})f=b.
\end{equation}

Inspired by \cref{W_ssl_for_wnll_definition}, we would like to define an affinity matrix that takes into account the known labels, such that it also distinguishes between labeled samples from the same and from different clusters.

\subsection{Semi-Supervised Laplacian Definition}
The classical definition of the graph-Laplacian, \cref{Laplacian_def}, is based on the data features in an unsupervised manner. In this section, we introduce a novel definition of the graph affinity matrix for SSL problems. That means, the aﬀinity measure takes into account not only the feature vectors but also the known information about the labels of a subset $S$ of $V$. 
The proposed definition is intended to improve performance for SSL clustering problems. 
For $K$ clusters, we denote by $S_k$ the set of labeled nodes belonging to the $k$-th cluster, such that
$S=\cup_{k=1}^KS_k$.

We suggest the following affinity measure

\begin{equation}\label{W_ssl definition}
W_{SSL}= 2W+\alpha W^{labeled},
\end{equation}
where $W$ is the known unsupervised affinity matrix, $\alpha$  is a scalar parameter, we set 
\begin{equation}
    \label{eq:alpha}
    \alpha = \mu-1,
\end{equation}
with $\mu$ as in \eqref{eq:mu}
and $W^{labeled}$ is defined as follows,
\begin{equation}\label{W_ssl_labeled_definition}  W^{labeled}_{ij}=
  \begin{cases} 
     \max(W)  & x_i,x_j \in S_k  \text{   ,   } \forall k \in \{1,..K\}\\
     -\frac{2}{\alpha}W_{ij} & x_i\in S_k, x_j \in S_l \text{   ,   } \forall k,l \in \{1,...,K | k \neq l\} \\
     W_{ij} & x_i \in S, x_j \in X \setminus S \text{ or }  x_i \in X \setminus S, x_j \in S\\
     0 & x_i,x_j \in X \setminus S
  \end{cases}
.
\end{equation}

It can be observed that according to this definition, the connection between labeled nodes belonging to the same cluster is given the highest weight ($\max(W)$ is the maximum value of the unsupervised affinity matrix). This strong affinity ensures the nodes are well connected inducing high smoothness of the spectral solution at these regions. Labeled nodes of different clusters are disconnected. 
This increases the separation of these nodes, reducing smoothness and avoiding unnecessary regularity between nodes belonging to separate clusters. Note that although in the second line of \eqref{W_ssl_labeled_definition} the weights are negative, the final respective weights of $W_{SLL}$ are all non-negative. In addition,  as in \cref{W_labeled_wnll}, we reinforce edges between labeled nodes and unlabeled nodes. 
Now we can define the SSL graph-Laplacian as follows,  
 \begin{equation}\label{L_SSL def}
 L_{SSL}=D_{SSL}-W_{SSL},
 \end{equation}
 where $W_{SSL}$ is defined in \cref{W_ssl definition} and $D_{SSL}$ is its associated degree matrix (see  \cref{eq:DegreeMat_def}).
In a similar manner to \cref{Graph Laplacian matrix form}, one can solve the following problem
\begin{equation}\label{eq:GL_SSL_def}
(M \circ L_{SSL})f=b.
\end{equation}

\section{Analysis of $L_{SSL}$}
In this section, we analyze the characteristics of $L_{SSL}$ for different scenarios. 
First, we analyze the influence of each of the components in \cref{W_ssl_labeled_definition} on the spectral embedding.
Let us define the following affinity matrices, 
\begin{equation}  W^{1}_{ij}=
  \begin{cases} 
     \max(W)  & x_i,x_j \in S_k  \text{   ,   } \forall k \in \{1,..K\}\\
     0 & \textrm{otherwise}
  \end{cases}
\end{equation}

\begin{equation}  W^{2}_{ij}=
  \begin{cases} 
     -\frac{2}{\alpha}W_{ij} & x_i\in S_k, x_j \in S_l \text{   ,   } \forall k,l \in \{1,..K | k \neq l\}\\
     0 & \textrm{otherwise}
  \end{cases}
\end{equation}

\begin{equation}  W^{3}_{ij}=
  \begin{cases} 
     W_{ij} & x_i \in S, x_j \in X \setminus S \text{ or }  x_i \in X \setminus S, x_j \in S\\
     0 &  \textrm{otherwise}
  \end{cases}
\end{equation}
Note that $W^1$ and $W^2$ can be interpreted as \emph{Contrastive Affinities}, following insights of contrastive learning \cite{hadsell2006dimensionality}. 
The contrastive paradigm aims at creating an embedding where instances of the same cluster are very close (the role of  $W^1$), whereas instances of different clusters are distinctly separated (the role of  $W^2$).
On the other hand, $W^3$ can be thought of as \emph{Density Affinity}. 
Its purpose is to increase the density of the graph in the vicinity of labeled nodes.

\begin{figure}
    \captionsetup[subfigure]{justification=centering}
    \centering
    \begin{subfigure}{0.48\textwidth}
        \centering
        \includegraphics[trim =  1mm 1mm 1mm 1mm,clip,width=\textwidth,valign=t]{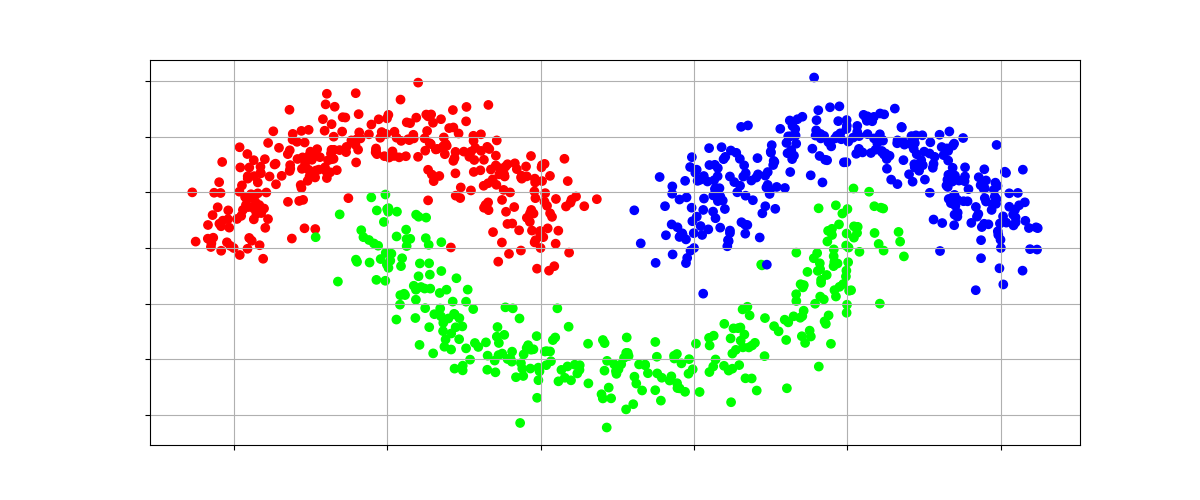}
        \caption{}
        \label{fig:li_3_moons_dataset}
    \end{subfigure}
    \hfill
    \begin{subfigure}{0.48\textwidth}
        \centering
        \includegraphics[trim =  1mm 1mm 1mm 1mm,clip,width=\textwidth,valign=t]{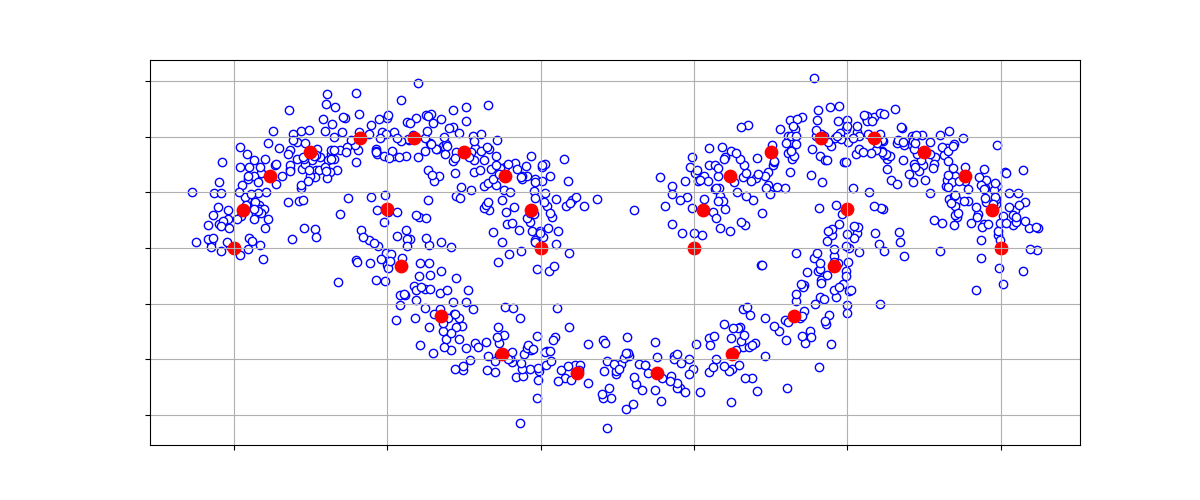}
        \caption{}
        \label{fig:li_labeled_nodes}
    \end{subfigure} 
    \caption{{\bf SSL Laplacian Illustration dataset.} The 3-Moons dataset containing 900 nodes appears in \cref{fig:li_3_moons_dataset}. The labeled nodes are shown in  \cref{fig:li_labeled_nodes}. }
    \label{fig:laplacian_illustration_dataset}
\end{figure}

We would like to use those affinity matrices instead of $ W^{labeled}$ in \cref{W_ssl definition} and examine the resulting spectral representation based on the graph-Laplacians $\{L_{SSL}^i\}_{i=1}^3$  defined by $\{W_{SSL}^i\}_{i=1}^3$  such that
\begin{equation}\label{W_ssl_i definition}
W^i_{SSL}= 2W+\alpha W^{i}.
\end{equation}
The spectral embedding is examined for the 3-Moons dataset containing $900$ nodes, of which $30$ are labeled, as can be seen in \cref{fig:laplacian_illustration_dataset}.

\begin{figure}
    \captionsetup[subfigure]{justification=centering}
    \centering
    \begin{subfigure}{0.19\textwidth}
        \centering
        \includegraphics[trim =  1mm 1mm 1mm 1mm,clip,width=\textwidth,valign=t]{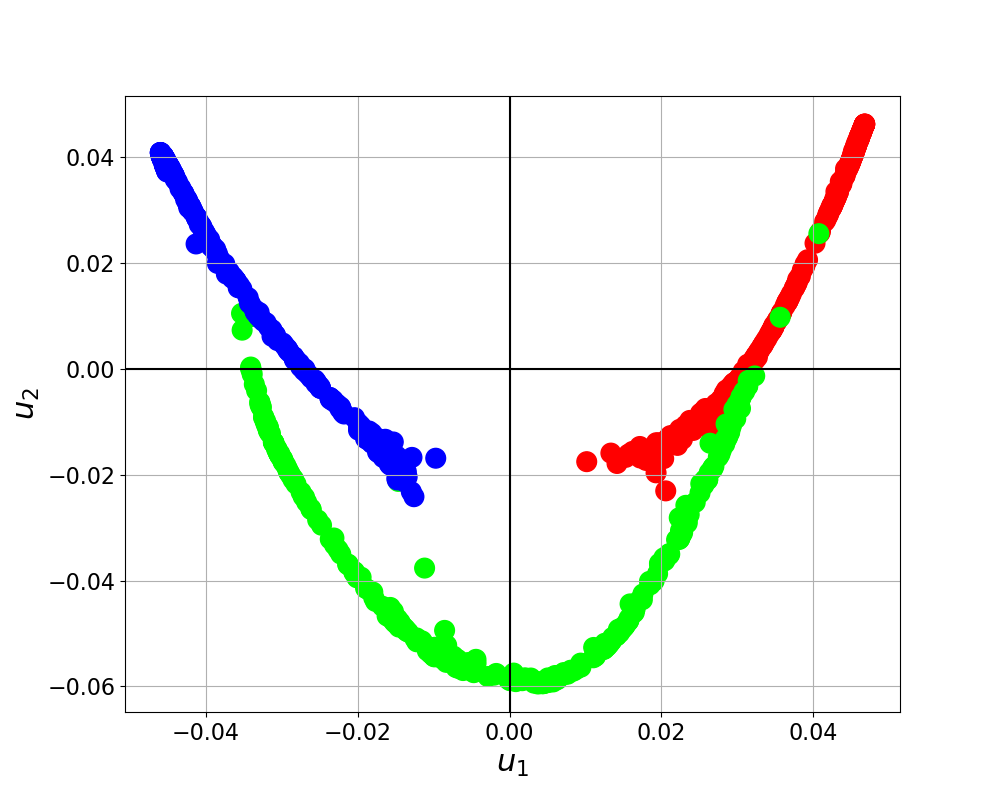}
        \caption{$L$}
        \label{fig:li_embd_us}
    \end{subfigure}
    \hfill
    \begin{subfigure}{0.19\textwidth}
        \centering
        \includegraphics[trim =  1mm 1mm 1mm 1mm,clip,width=\textwidth,valign=t]{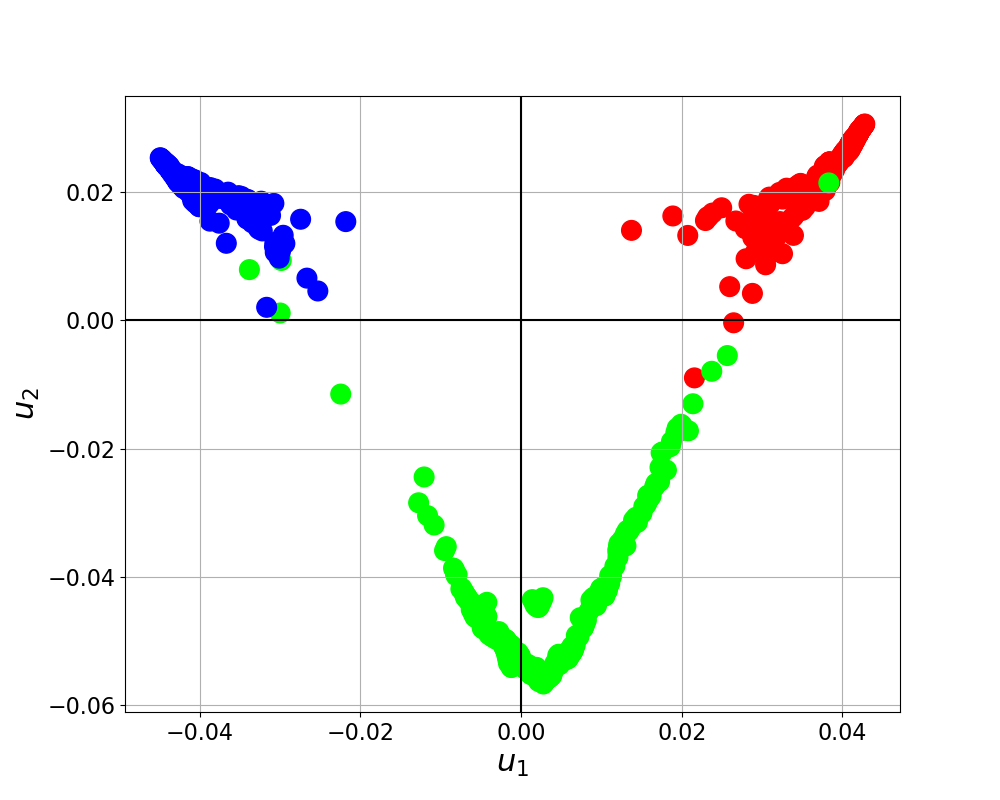}
        \caption{$L^1_{SSL}$}
        \label{fig:li_embd_maxonly}
    \end{subfigure} 
        \hfill
    \begin{subfigure}{0.19\textwidth}
        \centering
        \includegraphics[trim =  1mm 1mm 1mm 1mm,clip,width=\textwidth,valign=t]{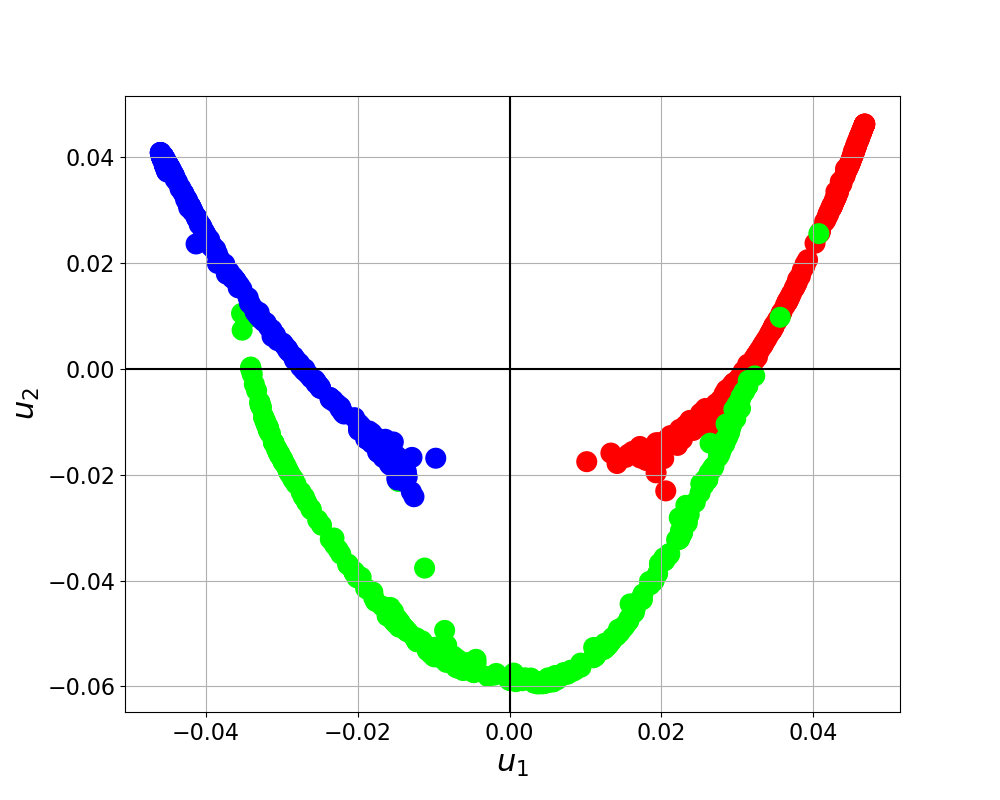}
        \caption{$L^2_{SSL}$}
        \label{fig:li_embd_disconnectonly}
    \end{subfigure} 
            \hfill
    \begin{subfigure}{0.19\textwidth}
        \centering
        \includegraphics[trim =  1mm 1mm 1mm 1mm,clip,width=\textwidth,valign=t]{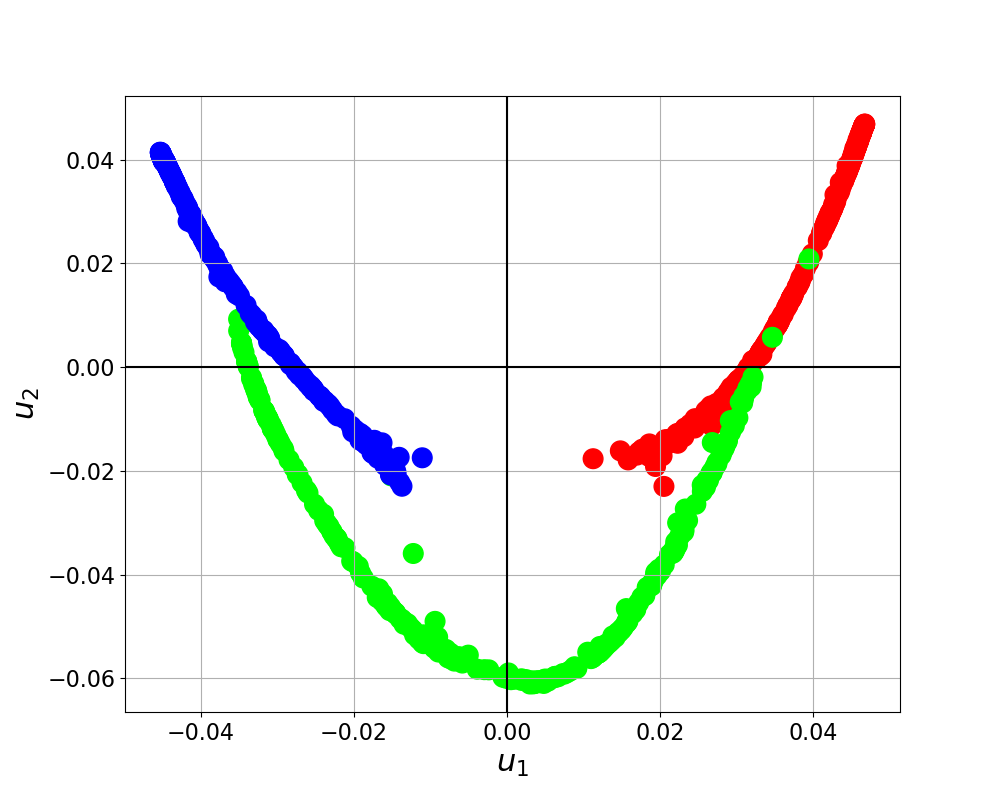}
        \caption{$L^3_{SSL}$}
        \label{fig:li_embd_wnll}
    \end{subfigure} 
                \hfill
    \begin{subfigure}{0.19\textwidth}
        \centering
        \includegraphics[trim =  1mm 1mm 1mm 1mm,clip,width=\textwidth,valign=t]{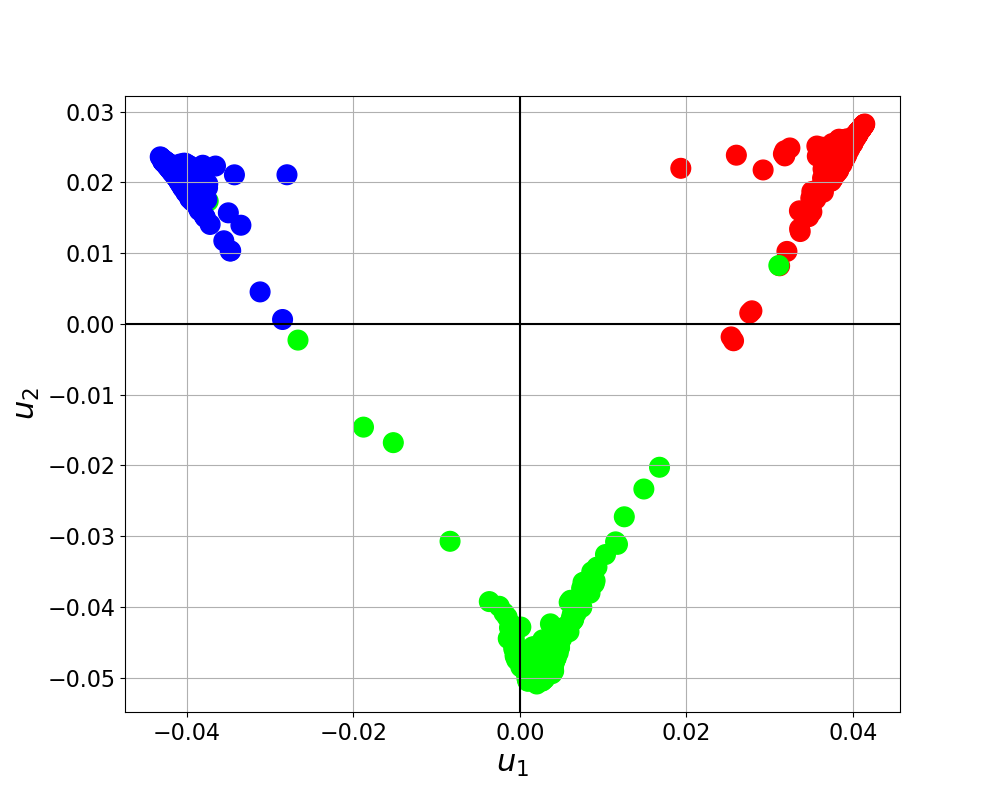}
        \caption{$L_{SSL}$}
        \label{fig:li_embd_ssl}
    \end{subfigure} 
    \caption{{\bf Spectral Embedding Illustration.} The spectral embedding obtained on the 3-Moons dataset for $L$, $\{L^i_{SSL}\}_{i=1}^3$ and $L_{SSL}$.}
    \label{fig:laplacian_illustration}
\end{figure}

We examine the spectral representation of the data, \cref{embedding_def}, spanned by the first two non-trivial leading eigenvectors of the unsupervised Laplacian $L$, $\{L^i_{SSL}\}_{i=1}^3$ and  $L_{SSL}$. The spectral embedding for each case is shown in \cref{fig:laplacian_illustration}. 
We can observe that the spectral representation obtained by $L_{SSL}$ produces the clearest division into clusters. Nodes of the same cluster (''moon'') are grouped together, whereas nodes of different clusters are further apart. An interesting finding in this experiment is that the main effect on the spectral embedding is caused by the contrastive affinities, especially of $W^1$. We will see in the experimental part this trend is valid also for more complex data. Indeed, the Laplacian based on $W^3$, which is equivalent to $L_{WNLL}$, has virtually no contribution to the spectral embedding.

\begin{figure}[H]
    \captionsetup{font={small}}
    \centering
    \columnname{$S$}\hfill
    \columnname{Spectral}\hfill
    \columnname{Spectral}\hfill
    \columnname{Dirichlet}\hfill
    \columnname{Dirichlet}\hfill
    \columnname{Dirichlet}\hfill
    \columnname{}\hfill
    \columnname{$L_{WNLL}$}\hfill
    \columnname{$L_{SSL}$}\hfill
    \columnname{$L$}\hfill
    \columnname{$L_{WNLL}$}\hfill
    \columnname{$L_{SSL}$}\\
    	    \begin{subfigure}{0.15\textwidth}
        \centering
        \includegraphics[trim =  1mm 1mm 1mm 1mm,clip,width=\textwidth,valign=t]{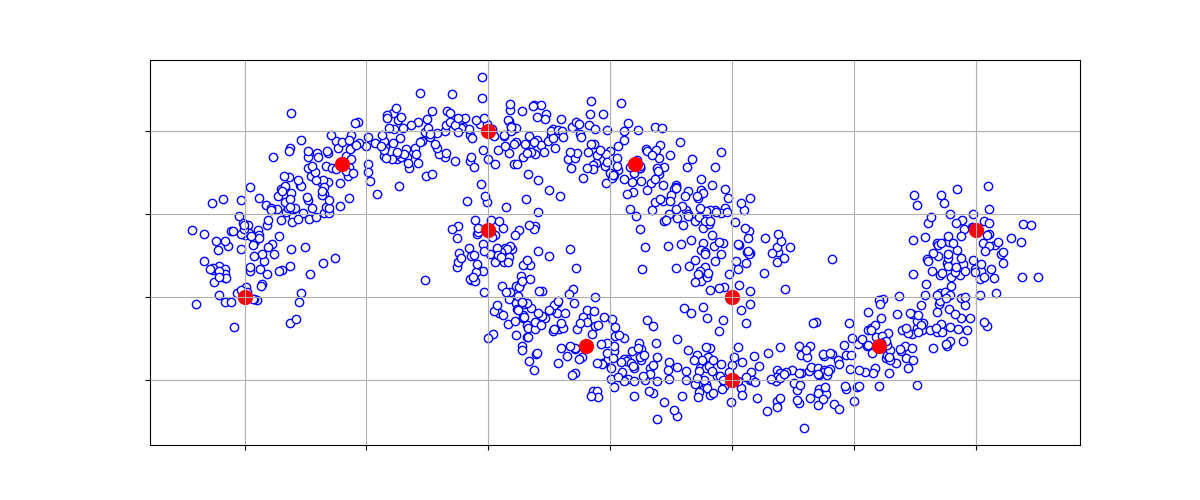}
        \caption*{$S_0$}
        \label{subfig:2moons_ssl_methods_noised_new3_labeled_option_7}
    \end{subfigure}
    \hfill
    \begin{subfigure}{0.15\textwidth}
        \centering
        \rotatebox[origin=c]{180}{\includegraphics[trim =  1mm 1mm 1mm 1mm,clip,width=\textwidth,valign=t]{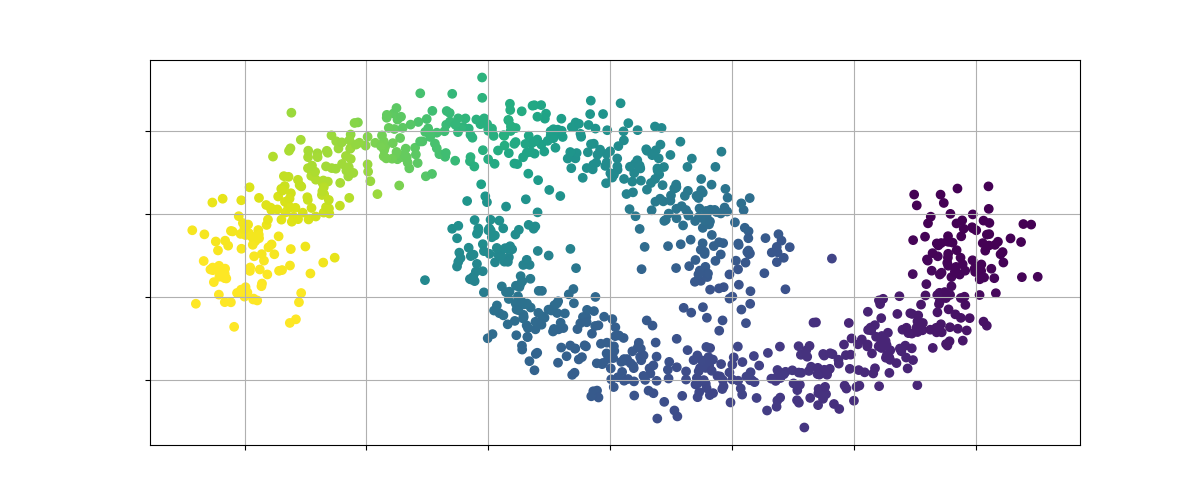}}
        \caption*{(0.42,0.80)}
        \label{subfig:2moons_ssl_methods_noised_new3_spectral_wnll_option_7_ev}
    \end{subfigure}
    \hfill
    \begin{subfigure}{0.15\textwidth}
        \centering
        \includegraphics[trim =  1mm 1mm 1mm 1mm,clip,width=\textwidth,valign=t]{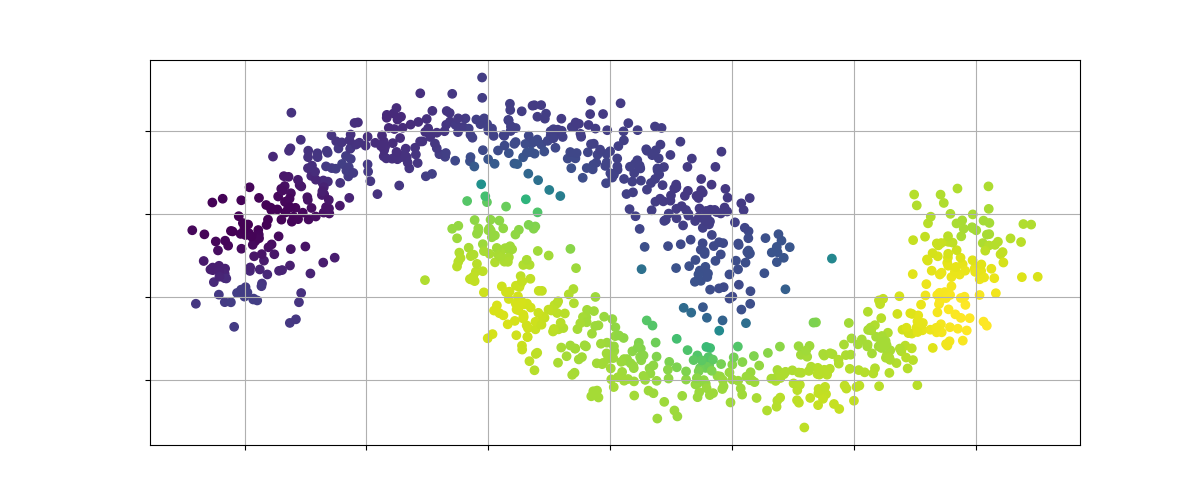}
        \caption*{(0.97,1.00)}
        \label{subfig:2moons_ssl_methods_noised_new3_spectral_ssl_option_7_ev}
    \end{subfigure}
    \hfill
    \begin{subfigure}{0.15\textwidth}
        \centering
        \includegraphics[trim =  1mm 1mm 1mm 1mm,clip,width=\textwidth,valign=t]{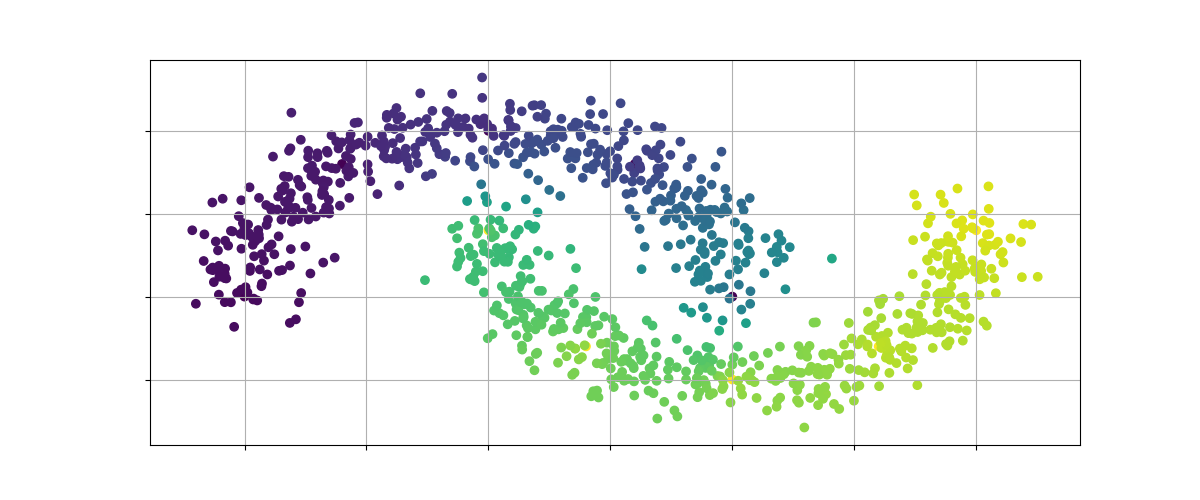}
        \caption*{(0.94,0.99)}
        \label{subfig:2moons_ssl_methods_noised_new3_dirichlet_US_option_7_phi}
    \end{subfigure}
    \hfill
    \begin{subfigure}{0.15\textwidth}
        \centering
        \includegraphics[trim =  1mm 1mm 1mm 1mm,clip,width=\textwidth,valign=t]{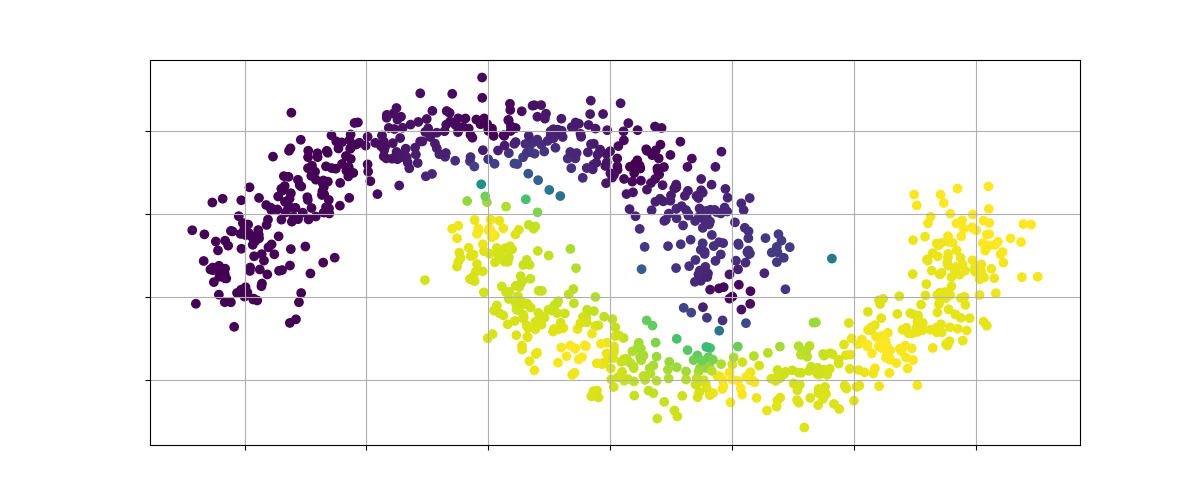}
        \caption*{(0.96,0.99)}
        \label{subfig:2moons_ssl_methods_noised_new3_dirichlet_WNLL_option_7_phi}
    \end{subfigure}
    \hfill
    \begin{subfigure}{0.15\textwidth}
        \centering
        \includegraphics[trim =  1mm 1mm 1mm 1mm,clip,width=\textwidth,valign=t]{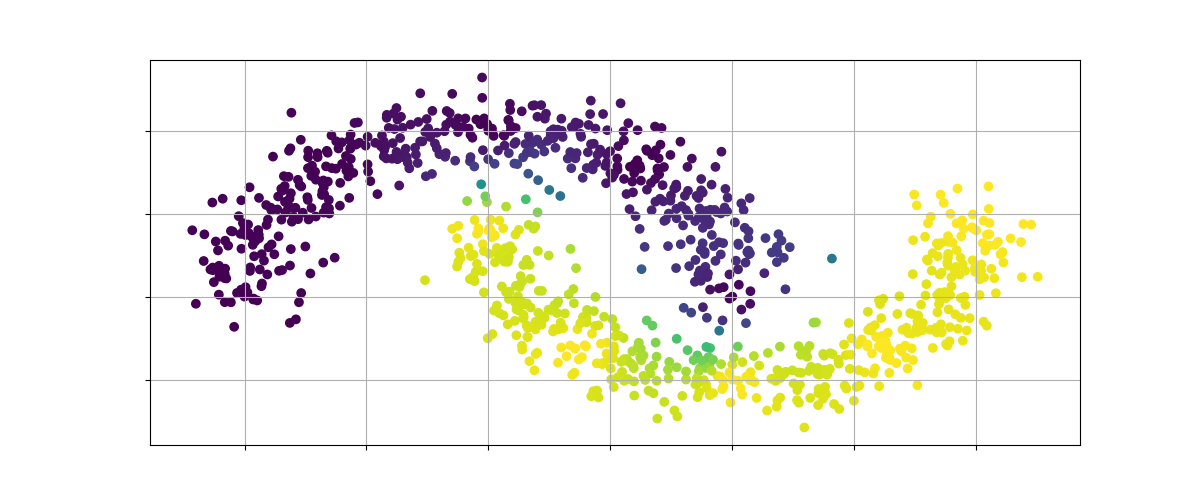}
        \label{subfig:2moons_ssl_methods_noised_new3_dirichlet_SSL_option_7_phi}
        \caption*{(0.96,0.99)}
    \end{subfigure}
    \hfill
	    \begin{subfigure}{0.15\textwidth}
        \centering
        \includegraphics[trim =  1mm 1mm 1mm 1mm,clip,width=\textwidth,valign=t]{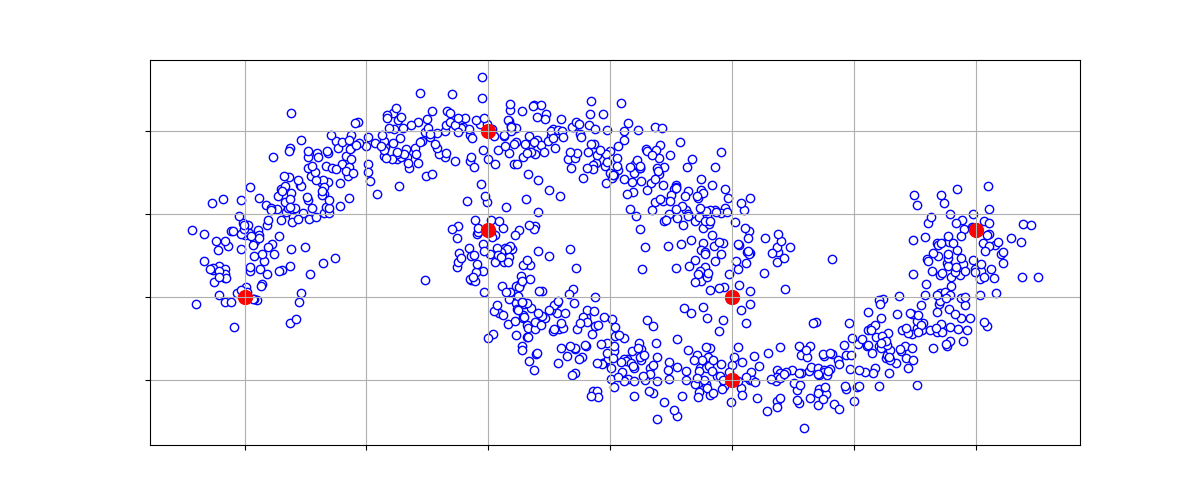}
        \caption*{$S_1$}
        \label{subfig:2moons_ssl_methods_noised_new3_labeled_option_0}
    \end{subfigure}
    \hfill
    \begin{subfigure}{0.15\textwidth}
        \centering
        \rotatebox[origin=c]{180}{\includegraphics[trim =  1mm 1mm 1mm 1mm,clip,width=\textwidth,valign=t]{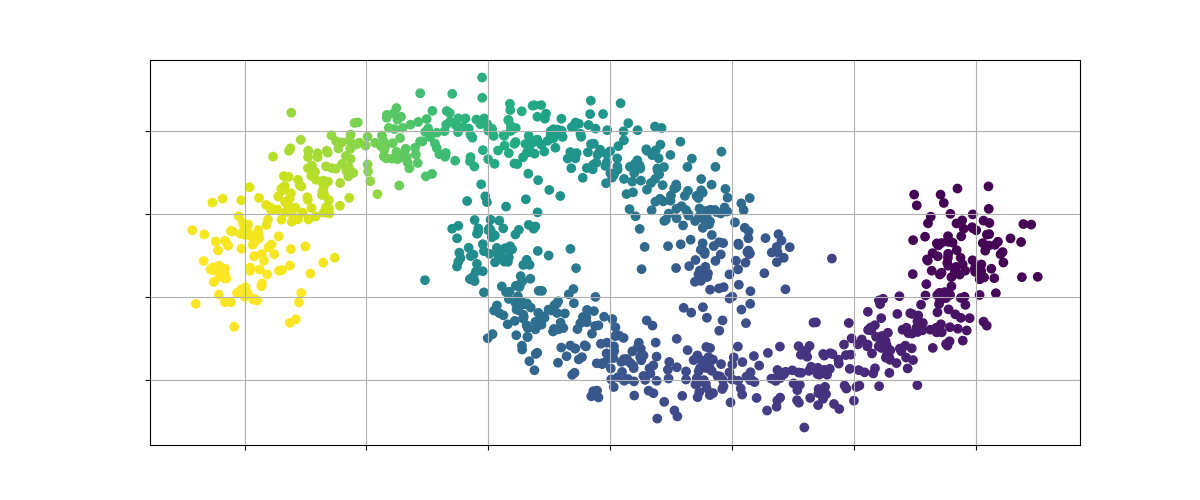}}
        \caption*{(0.43,0.81)}
        \label{subfig:2moons_ssl_methods_noised_new3_spectral_wnll_option_0_ev}
    \end{subfigure}
    \hfill
    \begin{subfigure}{0.15\textwidth}
        \centering
       \includegraphics[trim =  1mm 1mm 1mm 1mm,clip,width=\textwidth,valign=t]{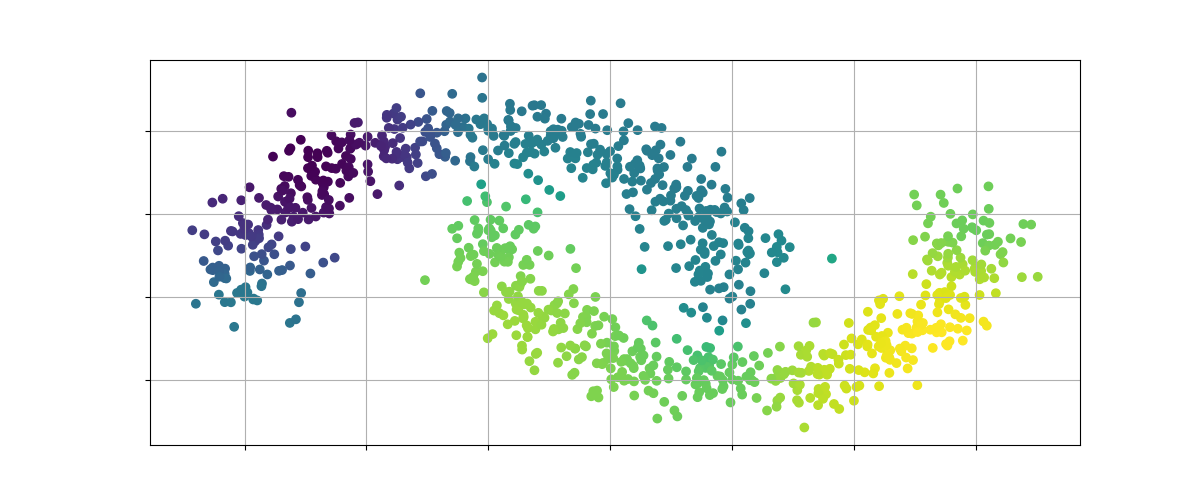}
        \caption*{(0.96,1.00)}
        \label{subfig:2moons_ssl_methods_noised_new3_spectral_ssl_option_0_ev}
    \end{subfigure}
    \hfill
    \begin{subfigure}{0.15\textwidth}
        \centering
        \includegraphics[trim =  1mm 1mm 1mm 1mm,clip,width=\textwidth,valign=t]{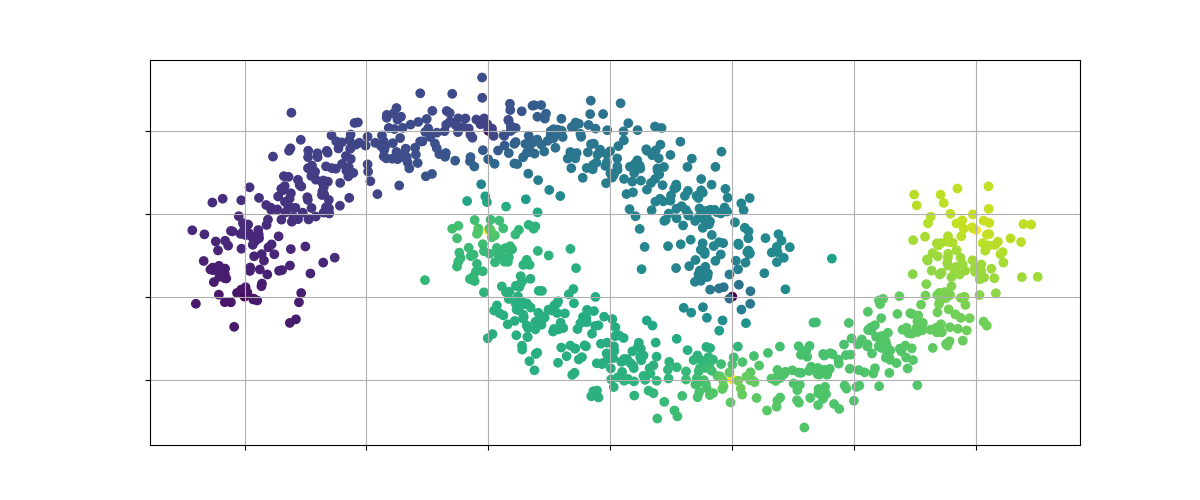}
        \caption*{(0.96,1.00)}
        \label{subfig:2moons_ssl_methods_noised_new3_dirichlet_US_option_0_phi}
    \end{subfigure}
    \hfill
    \begin{subfigure}{0.15\textwidth}
        \centering
        \includegraphics[trim =  1mm 1mm 1mm 1mm,clip,width=\textwidth,valign=t]{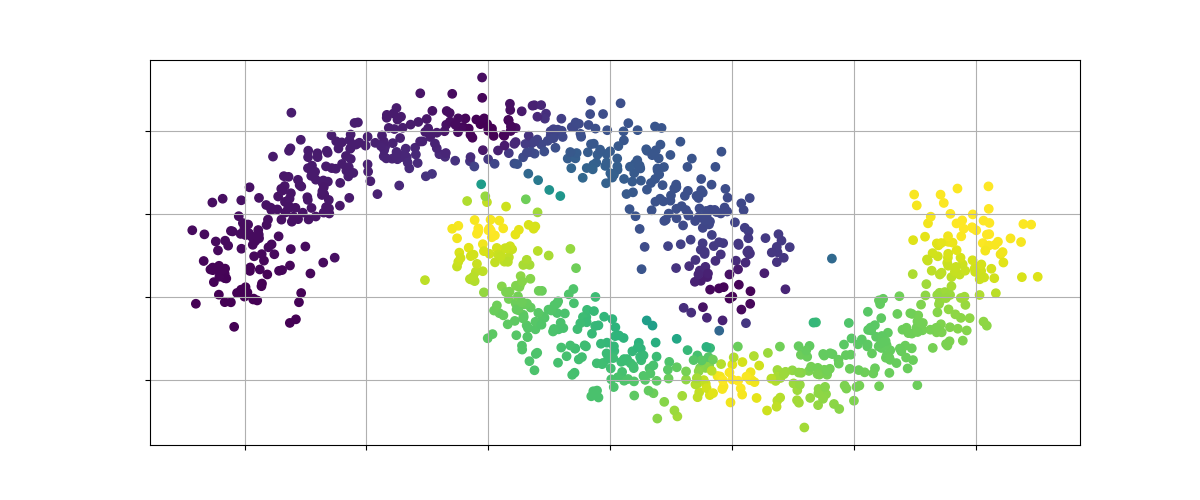}
        \caption*{(0.96,1.00)}
        \label{subfig:2moons_ssl_methods_noised_new3_dirichlet_WNLL_option_0_phi}
    \end{subfigure}
    \hfill
    \begin{subfigure}{0.15\textwidth}
        \centering
        \includegraphics[trim =  1mm 1mm 1mm 1mm,clip,width=\textwidth,valign=t]{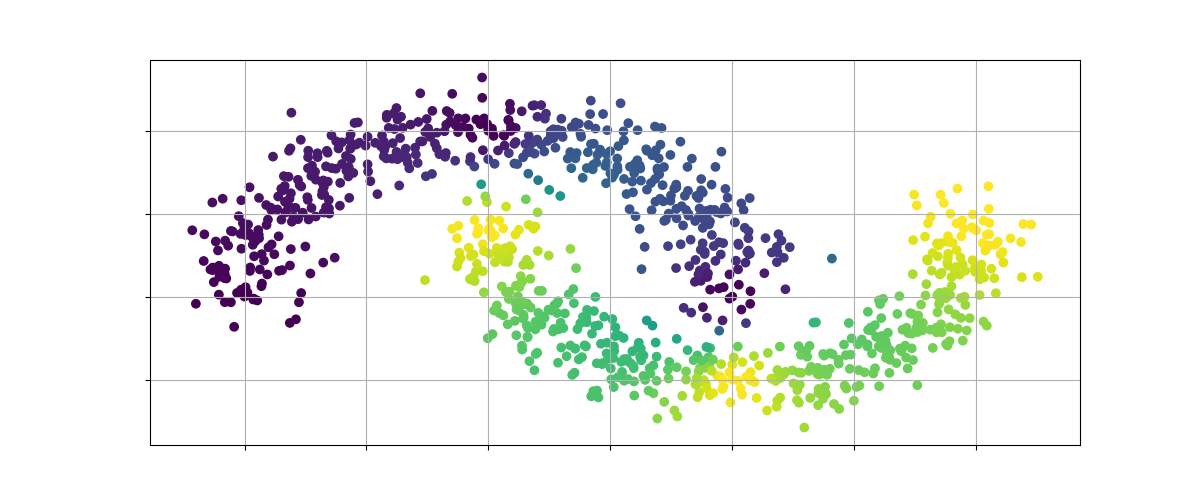}
        \label{subfig:2moons_ssl_methods_noised_new3_dirichlet_SSL_option_0_phi}
        \caption*{(0.96,1.00)}
    \end{subfigure}
    \hfill
	        \begin{subfigure}{0.15\textwidth}
        \centering
        \includegraphics[trim =  1mm 1mm 1mm 1mm,clip,width=\textwidth,valign=t]{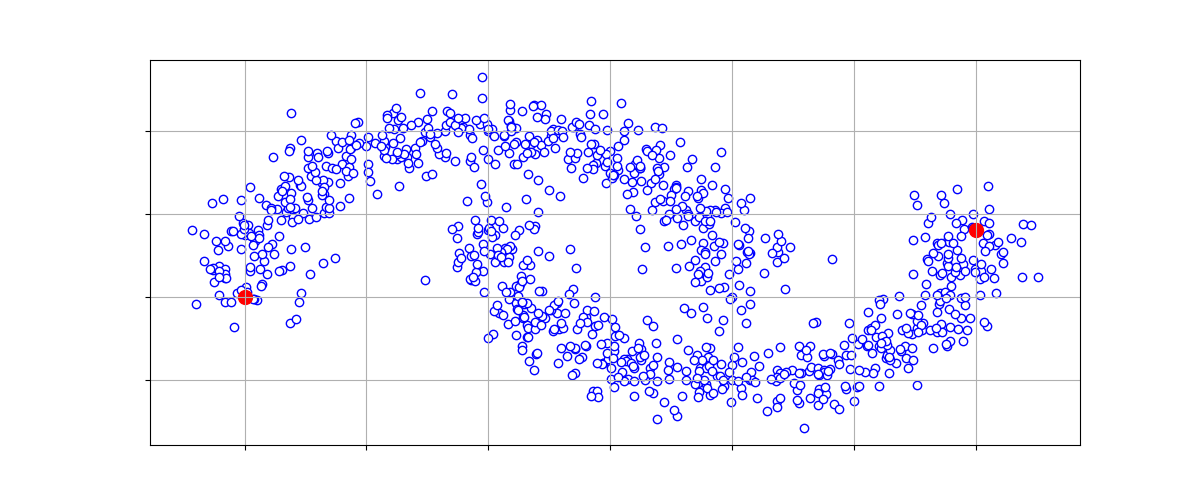}
        \caption*{$S_2$}
        \label{subfig:2moons_ssl_methods_noised_new3_labeled_option_1}
    \end{subfigure}
    \hfill
    \begin{subfigure}{0.15\textwidth}
        \centering
        \rotatebox[origin=c]{180}{\includegraphics[trim =  1mm 1mm 1mm 1mm,clip,width=\textwidth,valign=t]{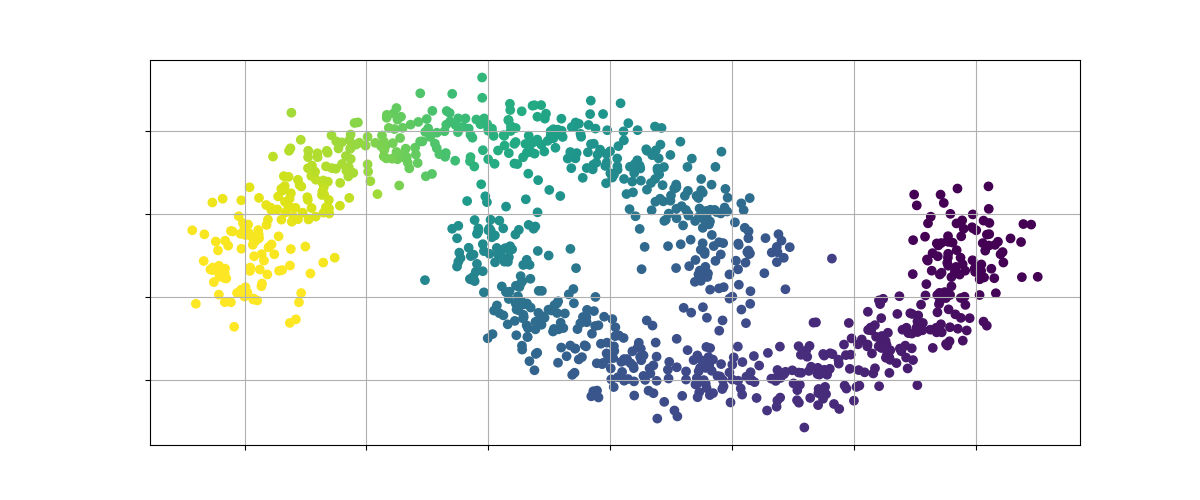}}
        \caption*{(0.42,0.81)}
        \label{subfig:2moons_ssl_methods_noised_new3_spectral_wnll_option_1_ev}
    \end{subfigure}
    \hfill
    \begin{subfigure}{0.15\textwidth}
        \centering
        \rotatebox[origin=c]{180}{\includegraphics[trim =  1mm 1mm 1mm 1mm,clip,width=\textwidth,valign=t]{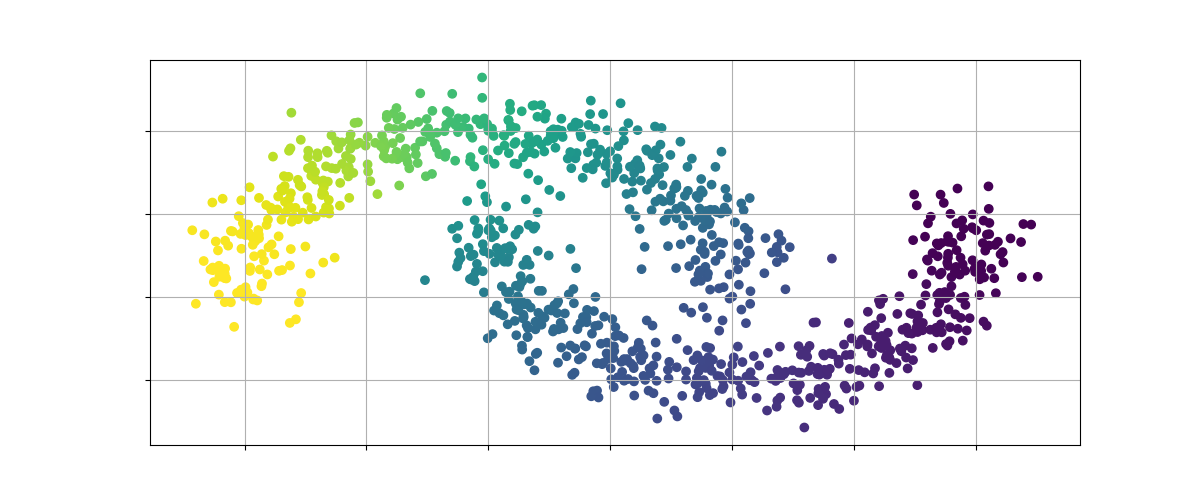}}
        \caption*{(0.42,0.81)}
        \label{subfig:2moons_ssl_methods_noised_new3_spectral_ssl_option_1_ev}
    \end{subfigure}
    \hfill
    \begin{subfigure}{0.15\textwidth}
        \centering
        \includegraphics[trim =  1mm 1mm 1mm 1mm,clip,width=\textwidth,valign=t]{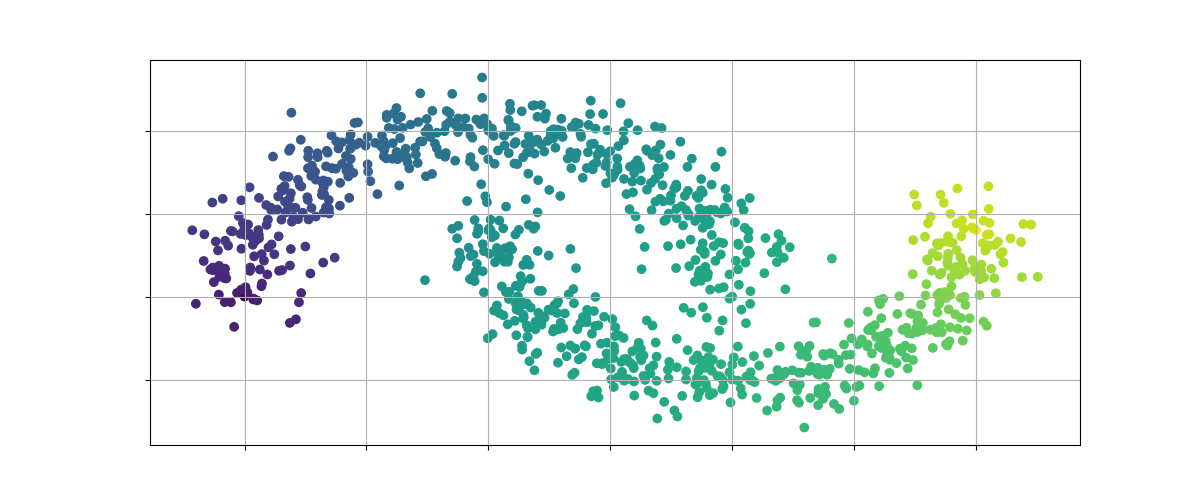}
        \caption*{(0.37,0.77)}
        \label{subfig:2moons_ssl_methods_noised_new3_dirichlet_US_option_1_phi}
    \end{subfigure}
    \hfill
    \begin{subfigure}{0.15\textwidth}
        \centering
        \includegraphics[trim =  1mm 1mm 1mm 1mm,clip,width=\textwidth,valign=t]{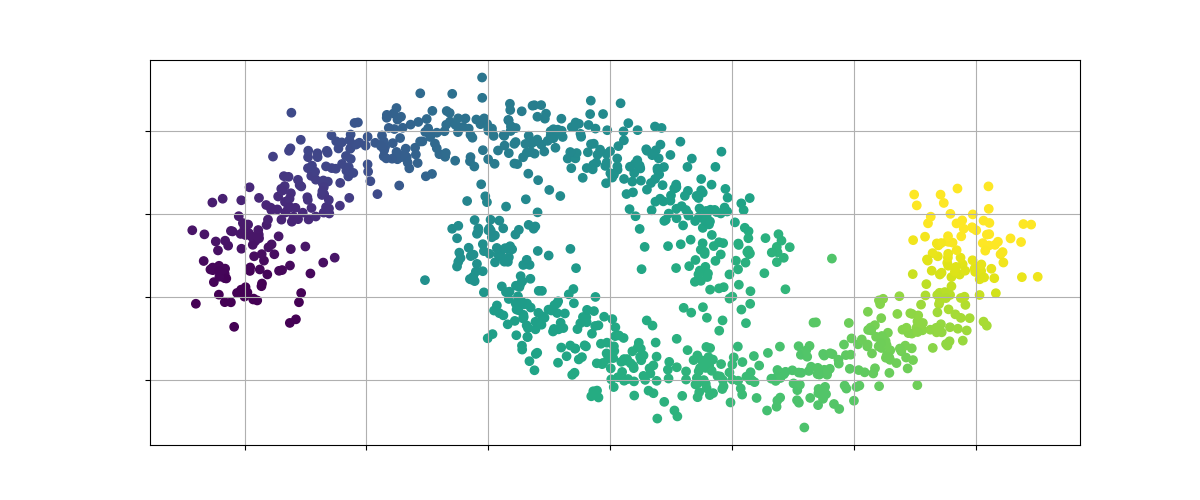}
        \caption*{(0.37,0.77)}
        \label{subfig:2moons_ssl_methods_noised_new3_dirichlet_WNLL_option_1_phi}
    \end{subfigure}
    \hfill
    \begin{subfigure}{0.15\textwidth}
        \centering
        \includegraphics[trim =  1mm 1mm 1mm 1mm,clip,width=\textwidth,valign=t]{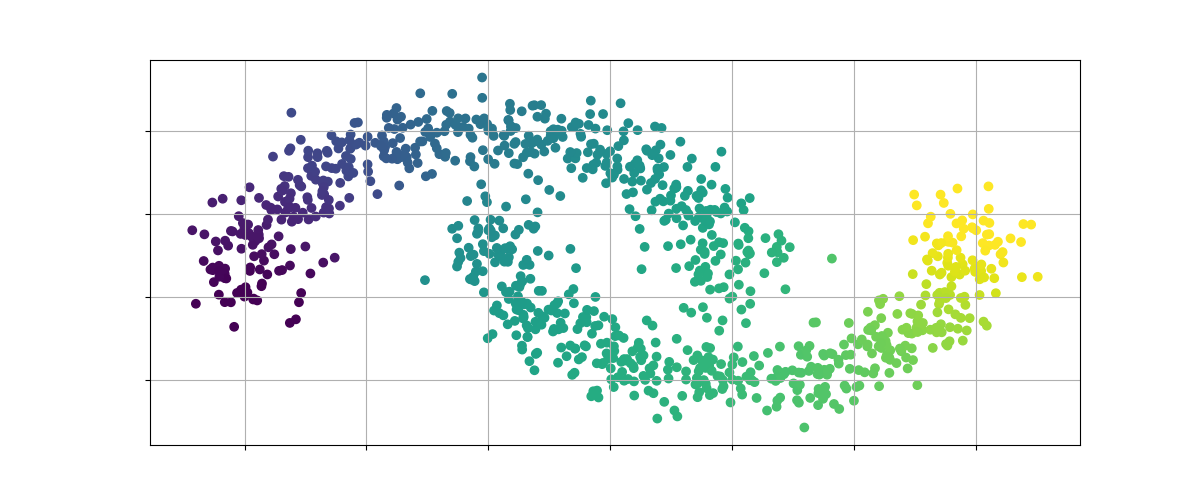}
        \label{subfig:2moons_ssl_methods_noised_new3_dirichlet_SSL_option_1_phi}
        \caption*{(0.37,0.77)}
    \end{subfigure}
    \hfill
	\begin{subfigure}{0.15\textwidth}
        \centering
        \includegraphics[trim =  1mm 1mm 1mm 1mm,clip,width=\textwidth,valign=t]{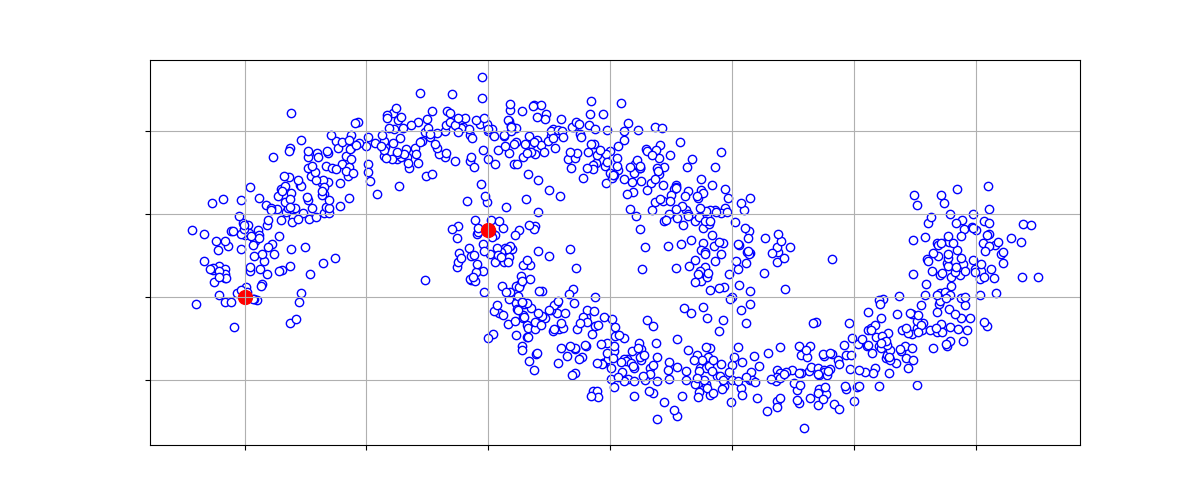}
        \caption*{$S_3$}
        \label{subfig:2moons_ssl_methods_noised_new3_labeled_option_4}
    \end{subfigure}
    \hfill
    \begin{subfigure}{0.15\textwidth}
        \centering
        \rotatebox[origin=c]{180}{\includegraphics[trim =  1mm 1mm 1mm 1mm,clip,width=\textwidth,valign=t]{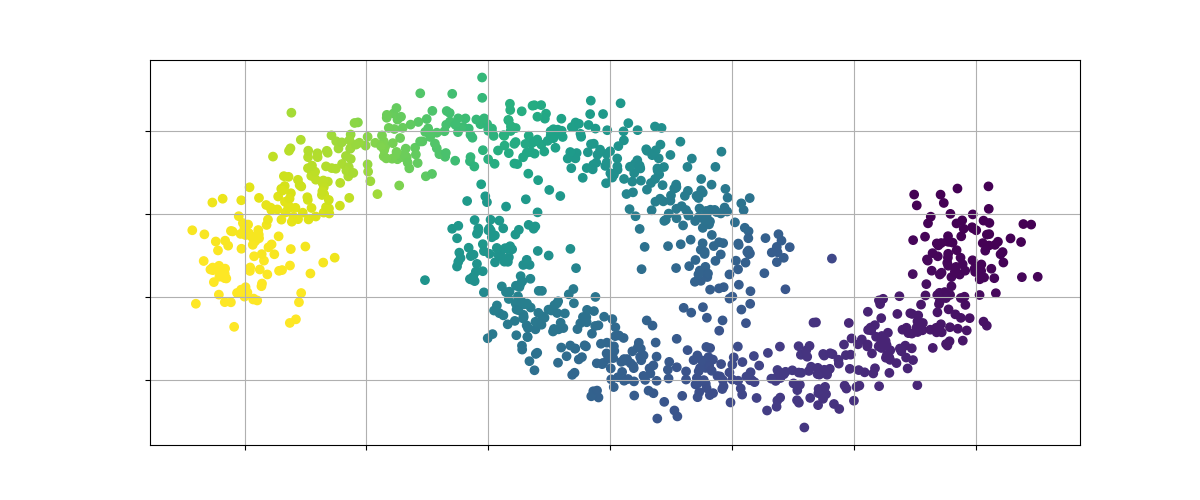}}
        \caption*{(0.43,0.81)}
        \label{subfig:2moons_ssl_methods_noised_new3_spectral_wnll_option_4_ev}
    \end{subfigure}
    \hfill
    \begin{subfigure}{0.15\textwidth}
        \centering
        \rotatebox[origin=c]{180}{\includegraphics[trim =  1mm 1mm 1mm 1mm,clip,width=\textwidth,valign=t]{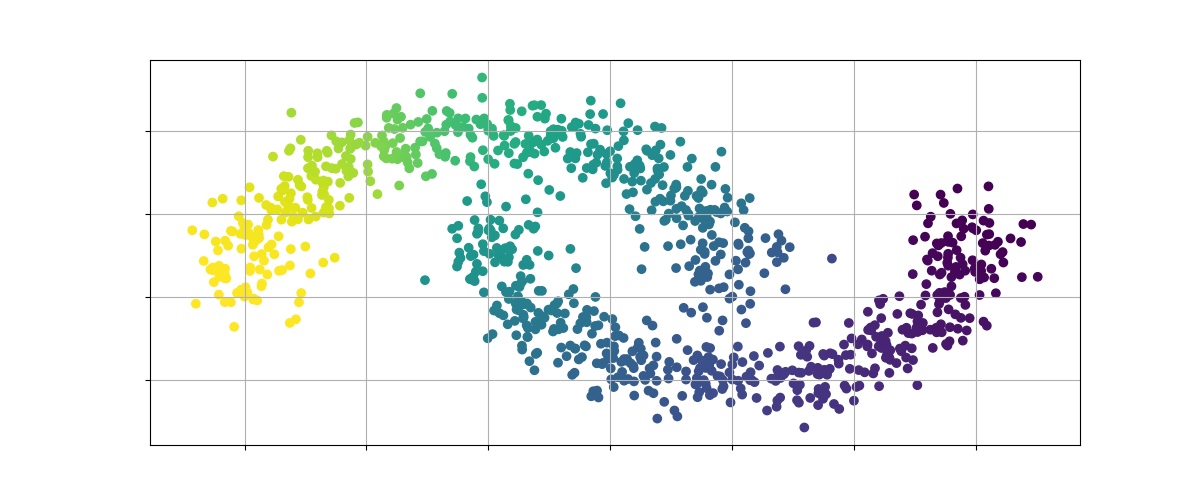}}
        \caption*{(0.43,0.81)}
        \label{subfig:2moons_ssl_methods_noised_new3_spectral_ssl_option_4_ev}
    \end{subfigure}
    \hfill
    \begin{subfigure}{0.15\textwidth}
        \centering
        \includegraphics[trim =  1mm 1mm 1mm 1mm,clip,width=\textwidth,valign=t]{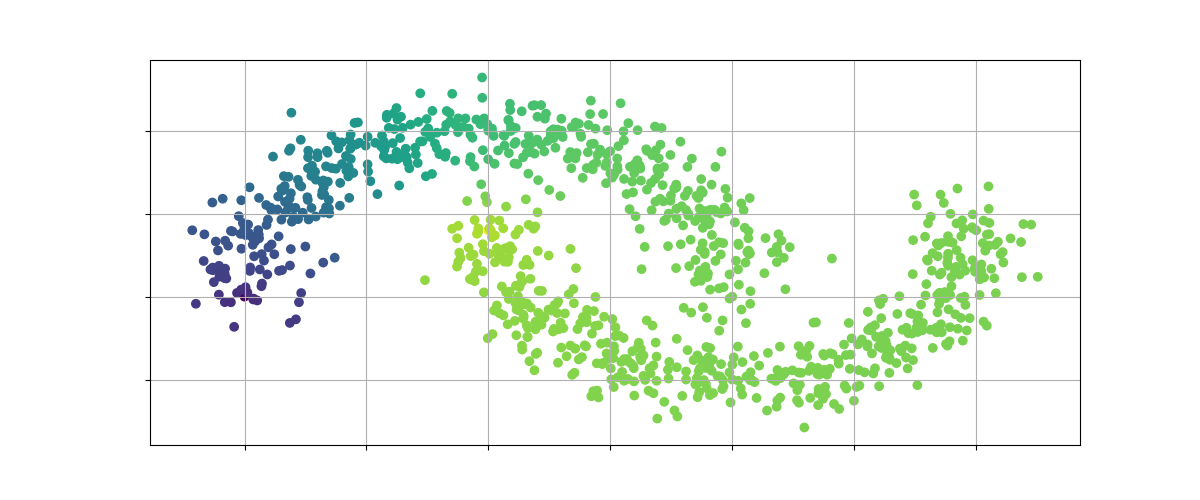}
        \caption*{(0.24,0.76)}
        \label{subfig:2moons_ssl_methods_noised_new3_dirichlet_US_option_4_phi}
    \end{subfigure}
    \hfill
    \begin{subfigure}{0.15\textwidth}
        \centering
        \includegraphics[trim =  1mm 1mm 1mm 1mm,clip,width=\textwidth,valign=t]{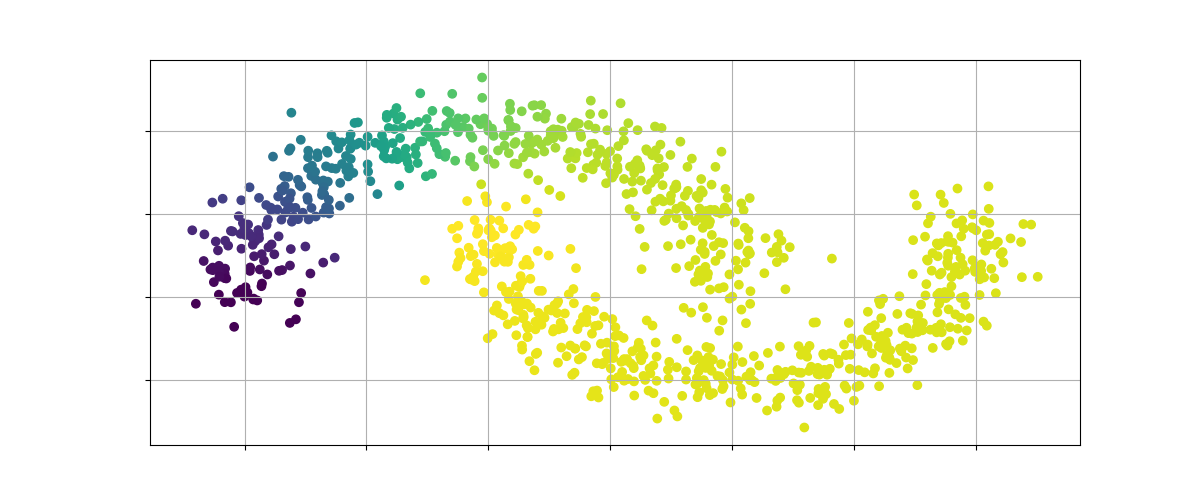}
        \caption*{(0.24,0.76)}
        \label{subfig:2moons_ssl_methods_noised_new3_dirichlet_WNLL_option_4_phi}
    \end{subfigure}
    \hfill
    \begin{subfigure}{0.15\textwidth}
        \centering
        \includegraphics[trim =  1mm 1mm 1mm 1mm,clip,width=\textwidth,valign=t]{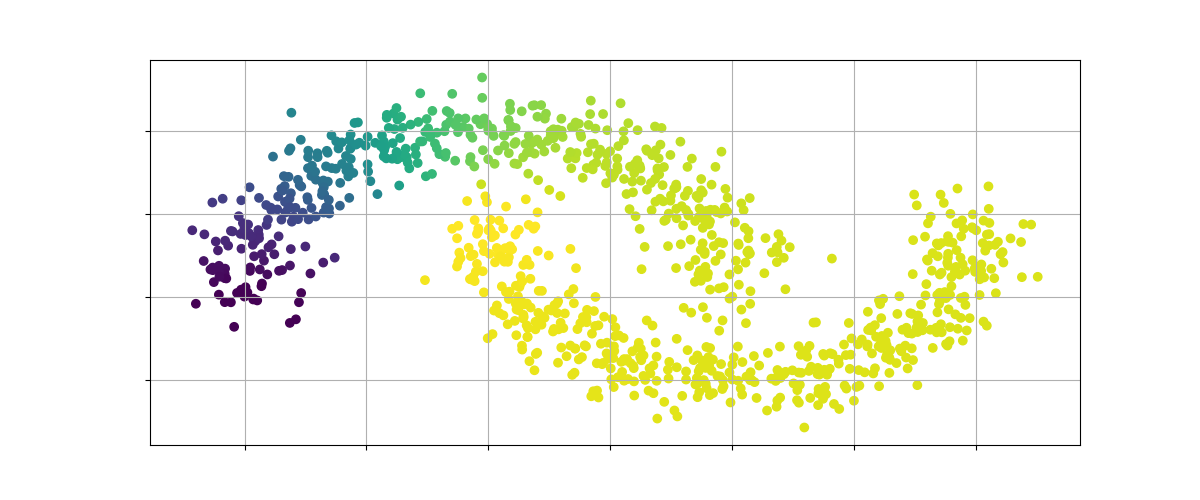}
        \label{subfig:2moons_ssl_methods_noised_new3_dirichlet_SSL_option_4_phi}
        \caption*{(0.24,0.76)}
    \end{subfigure}
    \caption{{\bf SSL solutions of the 2 Moons dataset.} 
    The first column shows different labeled sets $S$. The 2nd  and 3rd columns show the spectral clustering result for $L_{WNLL}$ and $L_{SSL}$, respectively. The 4th, 5th and 6th columns show the Dirichlet problem result for $L$,  $L_{WNLL}$ and $L_{SSL}$, respectively. The clustering measures (NMI, ACC) are presented below each figure.
    Spectral $L_{SSL}$ performs well in all configurations.
    } 
    \label{fig:2moons_ssl_methods_noised_new3}
\end{figure}

Next, we would like to examine spectral processing compared to constrained optimization, both based on $L_{SSL}$.
The performance of both approaches is tested over the 2-Moons dataset, which includes $1000$ instances. Each node of the graph is defined by its Euclidean position.
We analyze the spectral properties of the graph and the solution to the Dirichlet interpolation problem. The results are examined for different labeled sets $S$. 
In the spectral case, we find for each node of the graph its corresponding value according to the first non-trivial eigenvector of the graph-Laplacian, defined using $W_{WNLL}$ or $W_{SSL}$.  In order to find the solution for the Dirichlet problem, we set the value $1$ over the labeled nodes of the first moon and $-1$ over the labeled nodes of the second moon. Then we find the solution to the interpolation problem using $L$, $L_{WNLL}$ or $L_{SSL}$. 
To make a division into clusters, we perform K-Means over the resulting solution for each case. The obtained results are shown in \cref{fig:2moons_ssl_methods_noised_new3}.

Analyzing the results, it can be observed that when the amount of labeled samples is extremely small, the Dirichlet problem may not generalize that well the labels to the unlabeled data. This is especially true when the labels are not located near the cluster centers, as can be seen for the sets $S_2$ and $S_3$ (\cref{fig:2moons_ssl_methods_noised_new3} bottom two rows).
For the Dirichlet problem, the same performance is achieved for $L_{WNLL}$ and for $L_{SSL}$. This is also valid in larger data sets. 
In more complex scenarios the advantages of using the above Laplacians are clear, compared to standard $L$. In this toy example, the differences are minor. 
 
 Intermediate conclusions for these toy examples are that in some cases spectral analysis of the data is preferred. In addition, the suggested definition of $L_{SSL}$ allows us to get good performance for SSL problems both in the spectral case and for solving the Dirichlet problem. The reason for this is that $L_{SSL}$ includes the contrastive information, which is essential mainly for the spectral case, and the density information  which is more significant in the optimization case.
We will now test this in a more comprehensive manner.

\section{Experimental SSL Clustering Results}
In this section, we examine the performance of the different definitions of the graph-Laplacian for the clustering problem. To perform clustering in the semi-supervised case, we examine two different methods. The first one is {\bf Spectral Clustering} which is based on the division of the data into clusters by performing K-Means over the spectral embedding of the data, \cref{embedding_def}. The second method is based on {\bf Dirichlet-form Clustering}. To adapt the Dirichlet interpolation problem to multiple clusters, we use the algorithm suggested in \cite{shi2017weighted}.

\subsection{2-Moons Clustering} \label{sec:2_moons_clustering}
\begin{figure}[htb]
    \captionsetup[subfigure]{justification=centering}
    \centering
    \begin{subfigure}{0.24\textwidth}
        \centering
        \includegraphics[trim =  1mm 1mm 1mm 1mm,clip,width=\textwidth,valign=t]{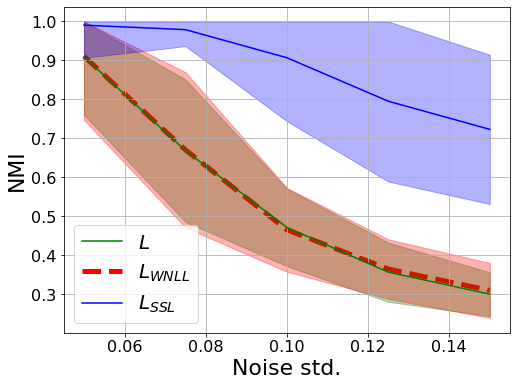}
        \caption{}
        \label{fig:nmi_vs_noise_L}
    \end{subfigure}
    \hfill
    \begin{subfigure}{0.24\textwidth}
        \centering
        \includegraphics[trim =  1mm 1mm 1mm 1mm,clip,width=\textwidth,valign=t]{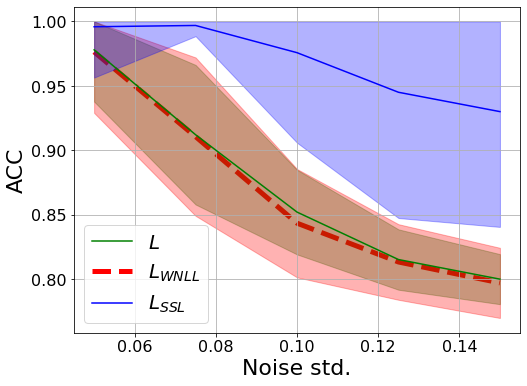}
        \caption{}
        \label{fig:acc_vs_noise_L}
    \end{subfigure} 
    \hfill
        \begin{subfigure}{0.24\textwidth}
        \centering
        \includegraphics[trim =  1mm 1mm 1mm 1mm,clip,width=\textwidth,valign=t]{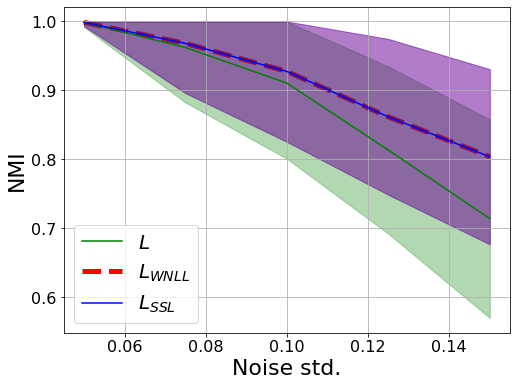}
        \caption{}
        \label{fig:nmi_vs_noise_GL}
    \end{subfigure}
    \hfill
    \begin{subfigure}{0.24\textwidth}
        \centering
        \includegraphics[trim =  1mm 1mm 1mm 1mm,clip,width=\textwidth,valign=t]{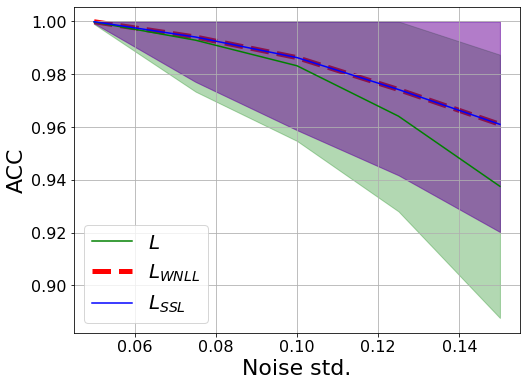}
        \caption{}
        \label{fig:acc_vs_noise_GL}
    \end{subfigure} 
    \caption{{\bf 2 Moons clustering for different noise parameter.} NMI and ACC measures over 100 different labeled set samples for different noise standard deviation. 
    Figs. \ref{fig:nmi_vs_noise_L}-\ref{fig:acc_vs_noise_L}
    are for Spectral Clustering.  Figs. \ref{fig:nmi_vs_noise_GL}-\ref{fig:acc_vs_noise_GL}
    are for Dirichlet Clustering. }
    \label{fig:rcut_ssl_vs_us_nmi_acc}
\end{figure}

\begin{figure}[h!]
    \captionsetup[subfigure]{justification=centering}
    \centering
    \begin{subfigure}{0.24\textwidth}
        \centering
        \includegraphics[trim =  1mm 1mm 1mm 1mm,clip,width=\textwidth,valign=t]{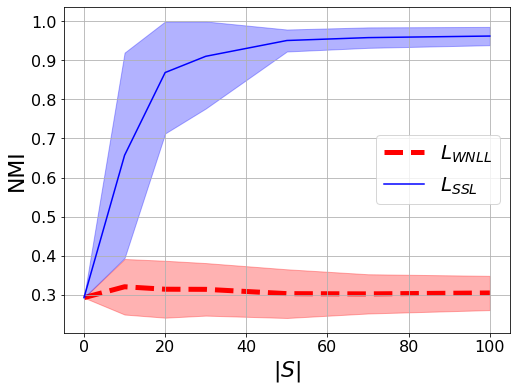}
        \caption{}
        \label{fig:nmi_vs_S_size_L}
    \end{subfigure}
    \hfill
    \begin{subfigure}{0.24\textwidth}
        \centering
        \includegraphics[trim =  1mm 1mm 1mm 1mm,clip,width=\textwidth,valign=t]{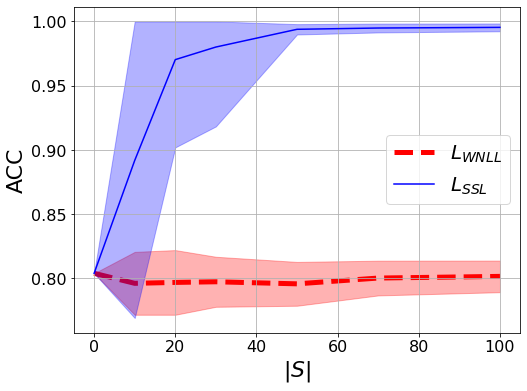}
        \caption{}
        \label{fig:acc_vs_S_size_L}
    \end{subfigure} 
    \hfill
        \begin{subfigure}{0.24\textwidth}
        \centering
        \includegraphics[trim =  1mm 1mm 1mm 1mm,clip,width=\textwidth,valign=t]{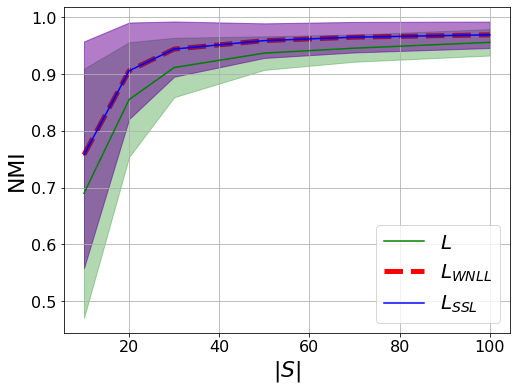}
        \caption{}
        \label{fig:nmi_vs_S_size_GL}
    \end{subfigure}
    \hfill
    \begin{subfigure}{0.24\textwidth}
        \centering
        \includegraphics[trim =  1mm 1mm 1mm 1mm,clip,width=\textwidth,valign=t]{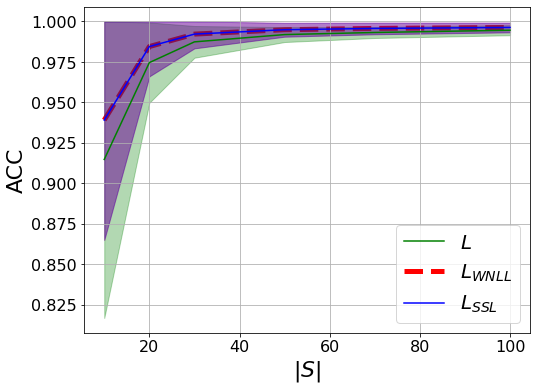}
        \caption{}
        \label{fig:acc_vs_S_size_GL}
    \end{subfigure} 
    \caption{{\bf 2 Moons clustering for different labeled set size.}  NMI and ACC measures over $100$ different labeled set samples for different labeled set sizes. Figs. \ref{fig:nmi_vs_S_size_L}-\ref{fig:acc_vs_S_size_L}
    are for Spectral Clustering.  Figs. \ref{fig:nmi_vs_S_size_GL}-\ref{fig:acc_vs_S_size_GL}
    are for Dirichlet Clustering.}
    \label{fig:rcut_S_size_nmi_acc}
\end{figure}

In this section, we present the statistical clustering performance for the 2-Moon dataset when the graph includes $500$ nodes. We perform two experiments. In the first one, shown in \cref{fig:rcut_ssl_vs_us_nmi_acc}, we examine the effect of changing the standard deviation of the noise of the data (that is, the deviation of each point from the position on the semicircle that defines the moon). In this case, $10$ labeled nodes from each class are randomly defined. In the second experiment, we examine the effect of changing the size of the labeled set $|S|$ (for fixed noise standard deviation set to $0.1$).
The results of this experiment are summarized in \cref{fig:rcut_S_size_nmi_acc}.
In both experiments, white Gaussian noise is used. The experiments show statistics of 100 trials, where a bold line represents the mean value (of NMI or ACC) and the lighter regions around each line depict the standard deviation of the measure.

The main conclusions from those experiments are that in the spectral case the results obtained for $L_{SSL}$  are much better compared to the other Laplacians, where $L_{WNLL}$ performance degenerates to the unsupervised case. For the Dirichlet problem, the performance of $L_{SSL}$ and $L_{WNLL}$ is similar and better than using the standard $L$, especially for the difficult cases, where the noise is significant and the amount of labeled information is small.
We can conclude that the definition of $L_{SSL}$ allows to get the best clustering performance when using the Dirichlet problem and especially for spectral analysis of the graph.

\begin{figure}[htb]
    \centering
    \includegraphics[width=0.40\textwidth]{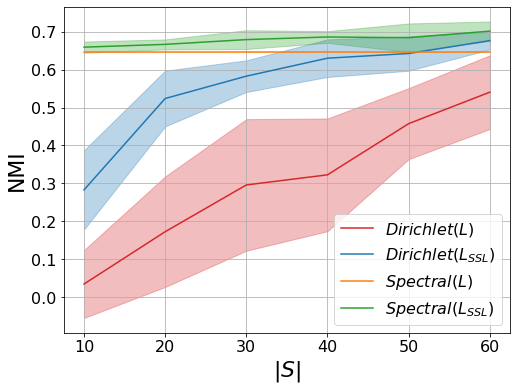}
    \includegraphics[width=0.40\textwidth]{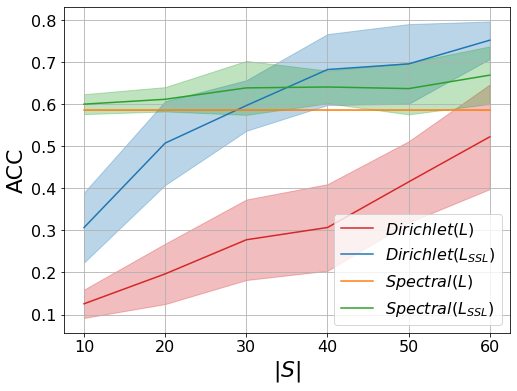}
    \includegraphics[width=0.40\textwidth]{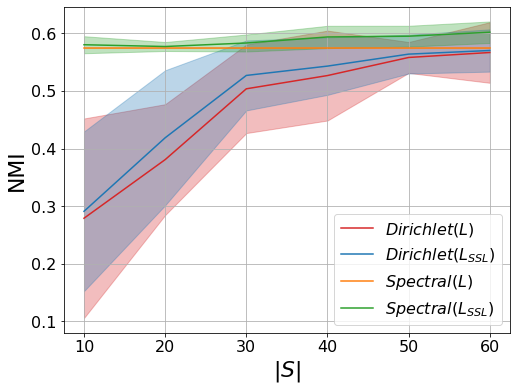}
    \includegraphics[width=0.40\textwidth]{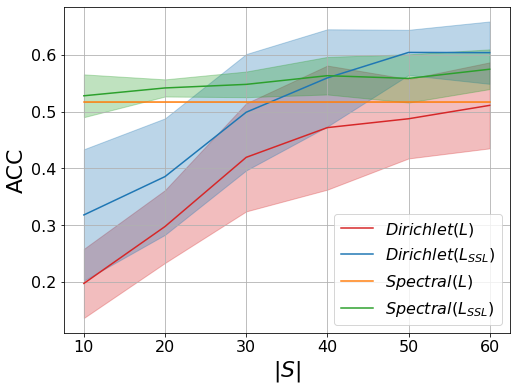}
    \caption{{\bf MNIST and F-MIST Clustering performance.} 
    The mean and standard deviation of NMI and ACC of 10 different experiments over the 10,000 samples of MNIST (Top row) and F-MNIST (bottom row) test sets, for different size of a labeled subset $|S|$.
    We note that $Spectral(L)$ has no variability since it does not depend on $S$. 
    }
    \label{fig:MNIST_and_FMNIST_all_nmi_acc}
\end{figure}

\subsection{MNIST and F-MNIST}
Now we examine the clustering performance over the MNIST and Fashion-MNIST datasets. Both of these well-known datasets include $28 \times 28$ gray-scale images. For each dataset, we define a graph using the test set which includes $10,000$ images. The obtained clustering performance, for different size of labeled sets, is shown in \cref{fig:MNIST_and_FMNIST_all_nmi_acc}.

It can be observed that using $L_{SSL}$ yields better performance, both for solving the Dirichlet problem and for spectral analysis of the graph. In addition, for small labeled set $|S|$, the performance obtained for spectral clustering is better.

\section{Conclusions}
In this paper, we propose a new definition for the graph-Laplacian designed to improve performance for SSL problems. 
The novel SSL Laplacian, which incorporates both contrastive and density affinities, yields improved spectral clustering and can be used also in constrained optimization problems. The proposed operator allows smooth interpolating between the unsupervised and the semi-supervised cases. The advantages are most prominent 
for an extremely low amount of labels or noisy data. 
In this work we have considered only the linear case, however, $p$-Laplacians may also be modified in a similar manner.
\subsection*{Acknowledgements}
We acknowledge support by grant agreement No. 777826 (NoMADS), by the Israel Science Foundation (Grant No.  534/19), by the Ministry of Science and Technology (Grant No. 5074/22) and by the Ollendorff Minerva Center.

\bibliographystyle{splncs04}
\bibliography{mybibliography}
\end{document}